\documentclass[sigconf, 10pt]{acmart}

\AtBeginDocument{%
  }

\usepackage{amsfonts}
\usepackage{amsmath}
\usepackage{booktabs}
\usepackage{makecell}
\usepackage{multirow}
\usepackage{subcaption}

\usepackage{microtype}
\setcopyright{acmlicensed}
\acmYear{2026}\copyrightyear{2026}
\setcopyright{cc}
\setcctype[4.0]{by}
\acmConference[MobiCom '26]{The 32nd Annual International Conference on Mobile Computing and Networking}{October 26--30, 2026}{Austin, TX, USA}
\acmBooktitle{The 32nd Annual International Conference on Mobile Computing and Networking (MobiCom '26), October 26--30, 2026, Austin, TX, USA}
\acmDOI{10.1145/3795866.3844469}
\acmISBN{979-8-4007-2505-0/26/10}

\acmSubmissionID{1241}

\begin{document}

\title[Group-Shared Low-Rank Approximation]{Group-Shared Low-Rank Approximation for Mobile-Efficient Pointwise Convolutions in Large-Kernel CNNs}

\author{
    Hao Luo$^1$,
    Yiting Yang$^1$,
    Wenyi Zhao$^1$,
    Man Jiang$^1$,
    Zhijun Lin$^2$, \\
    Ghulam Mohiuddin$^3$,
    Ting Jiang$^4$,
    Kunming Luo$^5$,
    Zihao Zhang$^6$,
    Qingsen Yan$^2$,
    Guoqing Wang$^7$,
    Wei Dong$^{1*}$,
    Peng Wang$^7$
}

\affiliation{
  \institution{
      $^1$College of Computer and Information Engineering, Xi'an University of Architecture and Technology
      $^2$Northwestern Polytechnical University
      \hspace{0.1cm}
      $^3$Nanchang University \\
      $^4$China Mobile Chengdu Institute of Research and Development \\
      $^5$Hong Kong University of Science and Technology \\
      $^6$Institute of AI for Industries, Chinese Academy of Sciences \\
      $^7$University of Electronic Science and Technology of China
  }
  \city{}
  \country{}
}

\renewcommand{\shortauthors}{Luo et al.}

\begin{abstract}
\renewcommand{\thefootnote}{\relax}
\footnotetext{* Wei Dong (dongwei156@xauat.edu.cn) is the corresponding author.}
\renewcommand{\thefootnote}{\arabic{footnote}}
Large-kernel Convolutional Neural Networks (CNNs) deliver remarkable performance in vision tasks by significantly expanding receptive fields, yet their quadratic parameter growth critically impedes storage-efficient edge deployment. While existing efficient architectures adopt parameter-efficient depthwise separable convolution backbones that leverage techniques like low-rank approximation and weight sharing to compress depthwise convolutions, we identify a critical oversight: pointwise convolutions dominate parameter volume ($>$87\% in models like RepLKNet-31B) and constitute the primary deployment bottleneck on resource-constrained edge devices. This results in prohibitive storage costs and severe memory-loading constraints on resource-limited devices (e.g., smartphones with 4-12 GB Random Access Memory (RAM)). To overcome this, we propose Channel Group-Shared (CGS) low-rank approximation, a novel Singular Value Decomposition (SVD)-based parameter-sharing strategy. CGS constructs a structured low-rank paradigm isomorphic to SVD decomposition, comprising shared (high-parameter-cost) down/up-projection matrices across channel groups within a layer and channel-group-specific (low-parameter-cost) scalable diagonal matrices. This group-sharing design achieves significant parameter reduction. Extensive experiments demonstrate that large-kernel CNNs (RepLKNet, ConvNeXt, SLaK) enhanced with CGS strike an empirically favorable balance between competitive performance and substantially reduced storage costs. Crucially, by alleviating storage constraints, reducing memory bandwidth pressure during loading, and minimizing model loading latency, CGS enables the feasible deployment of pre-trained large-kernel CNN models on edge devices, thereby bridging the gap between high-performance vision models and practical edge deployment.
\end{abstract}



\maketitle
\section{Introduction}

Large-kernel Convolutional Neural Networks~(CNNs) have recently emerged as a leading architecture for visual recognition tasks, owing to their ability to capture long-range dependencies through significantly expanded receptive fields~\cite{liu2022convnet, ding2022scaling, liu2022more}. The pursuit of expanding receptive fields has driven the development of large-kernel CNNs, which deliver remarkable performance in vision tasks such as image classification~\cite{dosovitskiy2020image, yuan2022volo}, object detection~\cite{yuan2022volo, liu2021swin, dai2021dynamic}, and semantic segmentation~\cite{xie2021segformer, wang2021pyramid, wang2021max, cheng2021per}. By leveraging convolutional kernels significantly larger than traditional $3\times 3$ filters (e.g., $7\times 7$ to $51\times 51$~\cite{liu2022convnet, ding2022scaling, liu2022more}), these models effectively capture long-range dependencies and contextual information, achieving State-of-the-Art (SOTA) results across a variety of applications. However, this performance gain comes at a steep cost: the quadratic growth of parameters within large kernels severely compromises storage efficiency. For instance, the number of parameters of a $31\times31$~\cite{ding2022scaling} kernel can reach approximately 100 times that of a 3×3 kernel.

To mitigate this issue, recent efficient large-kernel architectures (e.g., RepLKNet~\cite{ding2022scaling}, ConvNeXt~\cite{liu2022convnet}, SLaK~\cite{liu2022more}) predominantly adopt parameter-efficient depthwise separable convolution~\cite{chollet2017xception, howard2017mobilenets} as their backbone. This backbone includes two components: depthwise convolution and pointwise convolution. Depthwise convolution applies a single filter per input channel to capture spatial features, while pointwise convolution (a $1\times 1$ convolution) combines the output channels through linear projection. Together, they form the core of depthwise separable convolution, significantly reducing computation and parameters compared to standard convolution. 

\begin{figure}[tbph]
\centering
\includegraphics[width=\columnwidth]{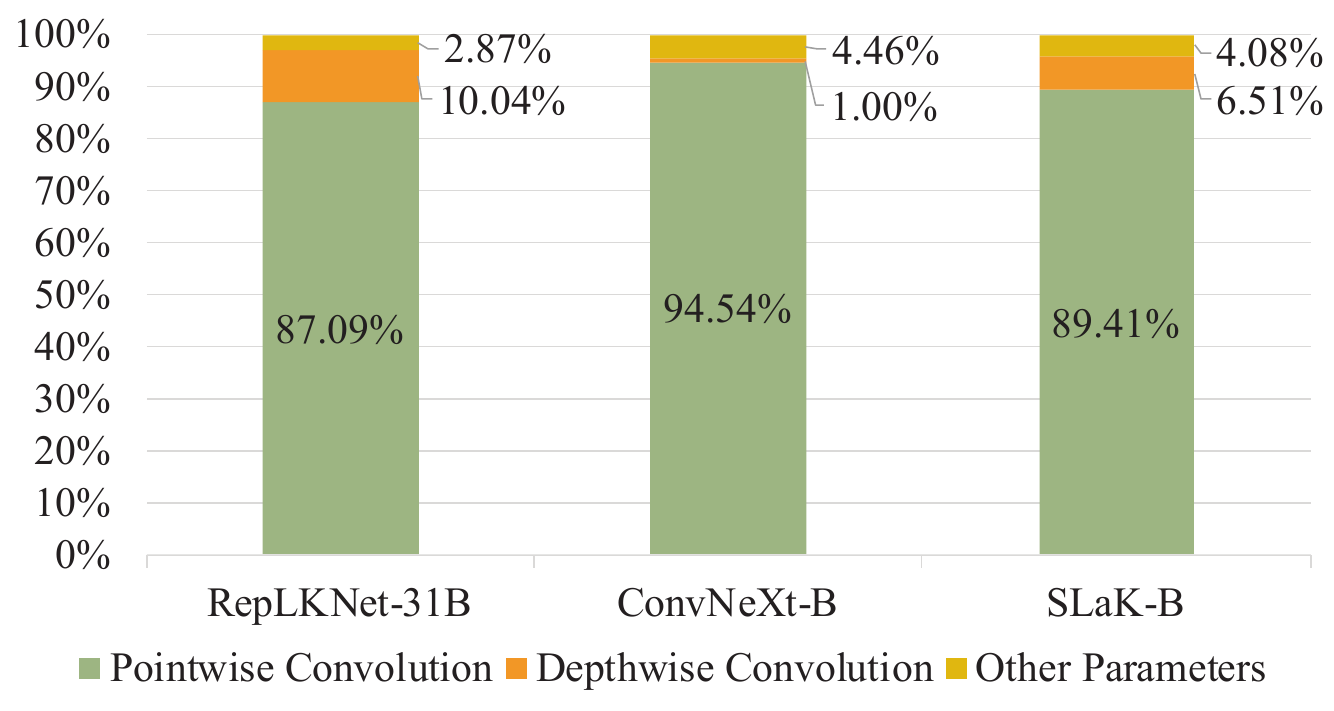}
\caption{Parameter distribution in large-kernel CNN models: proportions from pointwise convolutions, depthwise convolutions, and other components across RepLKNet-31B, ConvNeXt-B, and SLaK-B pre-trained models (where $B$ denotes the Base variant of each architecture).}
\label{fig:pw_model}
\end{figure}

\begin{figure}[t]
\centering
\includegraphics[width=\columnwidth]{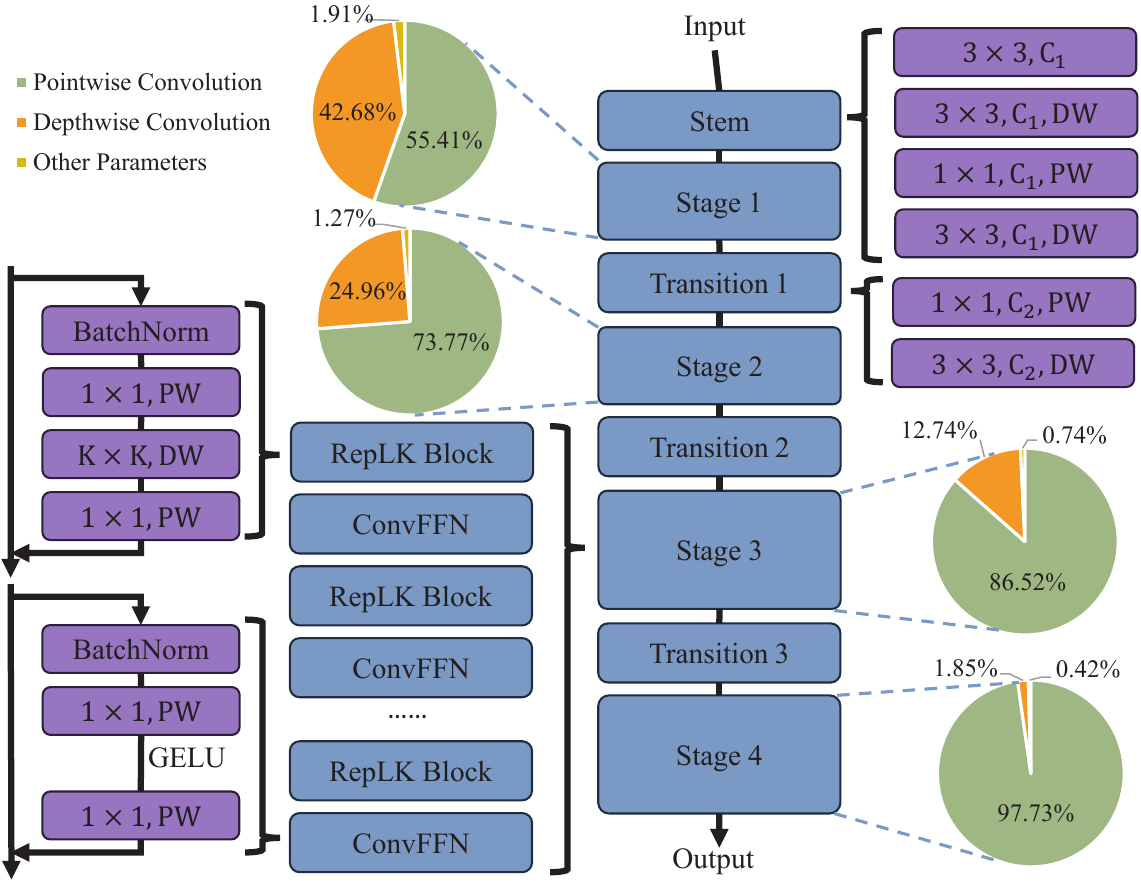}
\caption{Parameter distribution across stages in pre-trained RepLKNet-31B: pie charts quantify the proportion of parameters from pointwise convolutions, depthwise convolutions, and other components within each stage.}
\label{fig:pw_rep}
\end{figure}

\begin{figure}[t]
\centering
\includegraphics[width=\columnwidth]{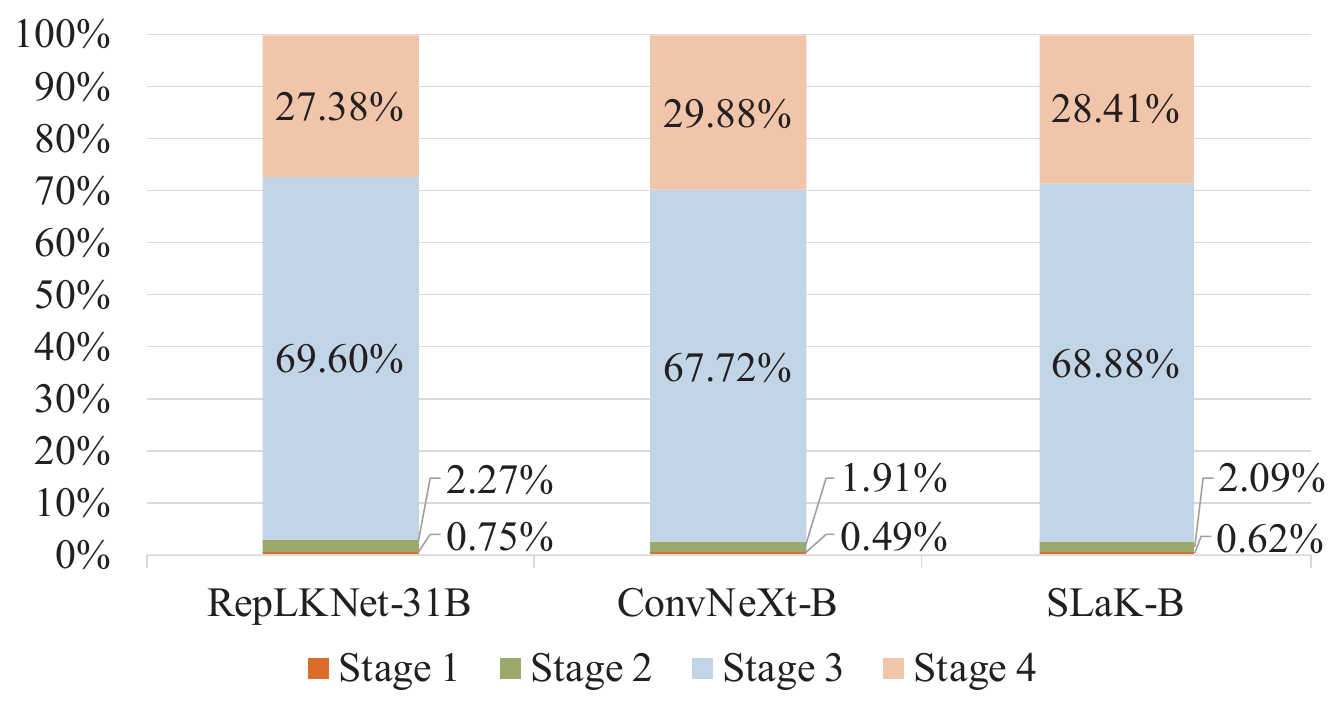}
\caption{Stage-wise parameter distribution across Stages 1–4 in pre-trained RepLKNet-31B, ConvNeXt-B, and SLaK-B models: each bar quantifies the proportion of parameters from individual stages relative to the total model parameters.}
\label{fig:pw_stage}
\end{figure}

Techniques like low-rank approximation~\cite{zhang2023speeding, tai2015convolutional}, weight sharing~\cite{chen2024pelk}, and dilated convolution~\cite{yu2017dilated, yu2015multi} have been successfully applied to compress parameters in the depthwise convolution component, achieving manageable model sizes for resource-constrained deployment. Yet, a critical bottleneck persists: pointwise convolutions now dominate the parameter volume of large-kernel CNN models, often accounting for over 87\% of the total parameters, as shown in Fig.~\ref{fig:pw_model} for the Base variants. This impedes the deployment of high-performance vision models on edge devices~\cite{sharma2022enabling,lai2018rethinking}~such as smartphones and Internet of Things (IoT) terminals, a critical step toward scalable and privacy-preserving artificial intelligence. Furthermore, a layer-wise analysis of RepLKNet-31B (Fig.~\ref{fig:pw_rep}) reveals that this parameter imbalance persists across all network stages, with pointwise convolutions consistently dominating parameter allocation. Concurrently, Fig.~\ref{fig:pw_stage} demonstrates that Stages 3 and 4 collectively account for approximately 97\% of total parameters in ImageNet-1K~\cite{deng2009imagenet} pre-trained RepLKNet-31B~\cite{ding2022scaling}, ConvNeXt-B~\cite{liu2022convnet}, and SLaK-B~\cite{liu2022more} architectures. This pronounced parameter concentration within later stages is characteristic of modern large-kernel CNNs. Surprisingly, despite its outsized impact, compressing pointwise convolutions in large-kernel CNNs remains largely unexplored; existing research focuses overwhelmingly on optimizing the depthwise convolution component while leaving this high-cost operator unaddressed. This greatly limits the process of deploying large-kernel CNN models on edge devices.

Therefore, deploying large-kernel CNNs~\cite{ding2022scaling} on edge devices~\cite{sharma2022enabling, lai2018rethinking} poses significant challenges, particularly \textbf{storage constraints} arising from massive parameter counts dominated by pointwise convolutions~\cite{hua2018pointwise, schuler2022grouped} that exceed device capacities and increase transmission overhead, and \textbf{runtime memory limitations} during model loading where redundant parameters cause inefficient memory bandwidth utilization while risking peak overload that may trigger system instability.

To bridge this gap, we propose Channel Group-Shared~(CGS) low-rank approximation, a novel parameter compression strategy grounded in Singular Value Decomposition (SVD) theory. Our key innovation lies in constructing a structured low-rank approximation paradigm that is isomorphic to the SVD factorized form. This paradigm decomposes a pointwise convolution layer into two components: (1)~High-parameter-cost down/up-projection matrices (analogous to left/right-unitary matrices in SVD) that are shared across multiple channel groups within the same layer; (2)~Low-parameter-cost group-specific scalable diagonal matrices (analogous to singular value diagonal matrix in SVD) whose elements are uniquely learned per channel group. By sharing the bulky projection matrices across groups while retaining lightweight group-specific scaling elements, CGS achieves dramatic parameter reduction without structural degradation. This makes the compressed large-kernel CNN models feasible for deployment on resource-constrained edge devices such as smartphones and IoT terminals. Specifically, the significant reduction in parameters directly alleviates the storage burden of edge devices, allowing the model to fit into limited local storage. Meanwhile, the streamlined parameter structure reduces the memory footprint during model loading and lowers model loading latency. In this way, CGS paves the way for deploying high-performance large-kernel convolution pre-trained models on edge devices, bridging the gap between SOTA model performance and real-world deployment constraints.

Our core contributions are fourfold:

\begin{itemize}
    \item Through empirical investigation, we identify that pointwise convolutions consume over 87\% of parameters in large-kernel CNNs, yet remain under-optimized. We thus propose an SVD-inspired CGS low-rank approximation that decomposes them into shared basis projections and group-adaptive scaling, significantly reducing parameters with lower capacity loss.
    \item We establish a mathematically-deduced compression framework derived from the isomorphism of SVD factorization, providing geometric interpretability and theoretical guarantees.
    \item Through extensive experiments on ImageNet-1K ~\cite{deng2009imagenet} classification, ADE20K~\cite{zhou2019semantic} semantic segmentation, and COCO~\cite{lin2014microsoft} object detection, we show CGS low-rank approximation reduces parameter by $>$81\% with $<$4.2\% accuracy drop (e.g., CGS-B versus RepLKNet-31B, as detailed in Tab.~\ref{tab:imagenet_results}), achieving an empirically observed balance between storage and accuracy among the compared compression methods.
    \item In terms of deployment on edge devices, we achieve the successful deployment of large-kernel convolutional pre-trained backbones on mobile platforms, and furthermore, empirically validate its effectiveness.
\end{itemize}
\section{Background and Motivation}

\subsection{Efficiency Challenges of Pointwise Convolutions in Large-Kernel CNNs}

Modern large-kernel CNNs (such as RepLKNet~\cite{ding2022scaling}, ConvNeXt~\cite{liu2022convnet}, SLaK~\cite{liu2022more}) have attained remarkable breakthroughs in visual tasks like image classification~\cite{dosovitskiy2020image, yuan2022volo}, object detection~\cite{yuan2022volo, liu2021swin, dai2021dynamic}, and semantic segmentation~\cite{xie2021segformer, wang2021pyramid, wang2021max, cheng2021per}, achieved through employing large-sized convolutional kernels ranging from $7\times7$~\cite{liu2022convnet} to $51\times51$~\cite{liu2022more}. By expanding the receptive field, these architectures effectively capture long-range dependencies, pushing the performance boundaries of a series of visual tasks. 

However, this architectural innovation also introduces significant storage efficiency challenges. Contrary to the common assumption that depthwise convolution are the primary driver of parameter growth, our analysis reveals that pointwise convolution constitute the dominant source of the overall parameter count. This highly skewed parameter distribution highlights a critical issue: while depthwise convolution are effective at capturing long-range spatial dependencies, the matrix multiplications of pointwise convolution, operating on high channel dimensions, have become the primary storage bottleneck, severely hindering model deployment on resource-constrained edge devices.

\vspace*{-0.23cm}

\subsection{Limitations of Existing Compression Methods}

Contemporary efficient architectures predominantly leverage depthwise separable convolution backbones to mitigate parameter explosion. Current research focuses extensively on depthwise convolution compression through techniques including: (1) \textbf{Kernel Factorization:} SLaK~\cite{liu2022more} decomposes oversized kernels (e.g., $51\times51 \rightarrow 51\times5 + 5\times51$) with dynamic sparsity; (2) \textbf{Parameter Sharing:} PeLK~\cite{chen2024pelk} employs a ``focus-blur" mechanism with exponential scaling grids for $>$90\% parameter reduction; (3) \textbf{Low-Rank Approximation:} convolutional kernels are compressed through low-rank approximation using Truncated SVD~\cite{zhang2023speeding, tai2015convolutional}; (4) \textbf{Dilated Convolutions:} multi-scale context aggregation without kernel expansion~\cite{yu2017dilated, yu2015multi}. 

Despite these advances, a critical limitation persists: existing methods universally neglect pointwise convolution, which dominates parameterization. This oversight creates a paradoxical efficiency gap, where compressing depthwise convolutions yields only marginal gains while pointwise operations remain prohibitively expensive for edge deployment. Specifically, (1) pointwise convolution parameters impose prohibitive storage demands; (2) memory loading requirements during initialization exceed mobile bandwidth capacities; and (3) in large-kernel CNNs, the compression of pointwise convolutions remains a bottleneck.
\vspace*{-0.23cm}
\section{Methodology}

This section first provides a brief background on Singular Value Decomposition (SVD)~\cite{lee2020parameter} and the equivalence properties of orthogonal matrices~\cite{garcia2018canonical, strang2012linear}. Based on this foundation, we then introduce the Channel Group-Shared (CGS) method for compressing pointwise convolutions by constructing kernels as a composition of shared down/up-projection matrices and channel-group-specific scalable diagonal matrices. Finally, we describe the deployment pipeline for transferring cloud pre-trained lightweight large-kernel convolutional foundation models to resource-constrained mobile devices for efficient execution.

\subsection{Preliminary Knowledge}

\subsubsection{Singular Value Decomposition Fundamentals}

Singular Value Decomposition (SVD)~\cite{lee2020parameter} provides a complete factorization of any real-valued matrix $\mathbf{W} \in \mathbb{R}^{m \times n}$ into orthogonal matrices and a diagonal matrix of singular values:
\begin{equation}
\mathbf{W} = \mathbf{U}\boldsymbol{\Sigma}\mathbf{V}^\top, 
\label{Eq:svd}
\end{equation}
where $\mathbf{U} \in \mathbb{R}^{m \times m}$ is the left singular matrix with orthogonal columns ($\mathbf{U}^\top\mathbf{U} = \mathbf{I}_m$); $\mathbf{V} \in \mathbb{R}^{n \times n}$ is the right singular matrix with orthogonal columns ($\mathbf{V}^\top\mathbf{V} = \mathbf{I}_n$); $\boldsymbol{\Sigma} \in \mathbb{R}^{m \times n}$ is the singular value matrix with diagonal elements $\sigma_1 \geq \sigma_2 \geq \cdots \geq \sigma_r > 0$ ($r = \operatorname{rank}(\mathbf{W})$) and $\Sigma_{ij} = 0$ for $i \neq j$.

A key property of this decomposition is its dimensional consistency. Any two matrices $\mathbf{W}_i, \mathbf{W}_j \in \mathbb{R}^{m \times n}$ of the same dimensions decompose into factors of identical shapes:
\begin{equation}
\begin{split}
\mathbf{W}_i &= \mathbf{U}_i \boldsymbol{\Sigma}_i \mathbf{V}_i^\top, \\
\mathbf{W}_j &= \mathbf{U}_j \boldsymbol{\Sigma}_j \mathbf{V}_j^\top,
\end{split}
\label{Eq:svd_same}
\end{equation}
where $\mathbf{U}_i, \mathbf{U}_j \in \mathbb{R}^{m \times m}$ and $\mathbf{V}_i, \mathbf{V}_j \in \mathbb{R}^{n \times n}$. Because the SVD factors share the same dimensions, their parameters can be structured and shared efficiently. This dimensional matching is the foundational insight for our compression framework.

\subsubsection{Orthogonal Matrix Equivalence}

Building on the orthogonal properties of SVD factor matrices in the real domain, we leverage a fundamental equivalence: any two orthogonal matrices of identical dimensions can be expressed as linear transformations of one another through invertible mappings~\cite{garcia2018canonical, strang2012linear}. Formally, for orthogonal matrices $\mathbf{M}_{\text{orth}_a}, \mathbf{M}_{\text{orth}_b} \in \mathbb{R}^{m \times m}$, there exist invertible matrices $\mathbf{P}, \mathbf{Q} \in \mathbb{R}^{m \times m}$ such that:
\begin{equation}
\mathbf{M}_{\text{orth}_a} = \mathbf{P}\mathbf{M}_{\text{orth}_b}\mathbf{Q}.
\label{Eq:orthogonal_transform}
\end{equation}

This equivalence enables our group-sharing paradigm. Consider $k$ orthogonal matrices $\{ \mathbf{U}_g \in \mathbb{R}^{m \times m} \}_{g=1}^k$ with identical dimensions. We express each through a shared base matrix $\mathbf{U}_{\text{shared}} \in \mathbb{R}^{m \times m}$ and group-specific transformations:
\begin{equation}
\left\{
\begin{array}{c}
\mathbf{U}_1 = \mathbf{P}_1\mathbf{U}_{\text{shared}}\mathbf{Q}_1 \\
\mathbf{U}_2 = \mathbf{P}_2\mathbf{U}_{\text{shared}}\mathbf{Q}_2 \\
\vdots \\
\mathbf{U}_k = \mathbf{P}_k\mathbf{U}_{\text{shared}}\mathbf{Q}_k
\end{array},
\right.
\label{Eq:group_sharing}
\end{equation}
where $\mathbf{P}_g, \mathbf{Q}_g \in \mathbb{R}^{m \times m}$ are invertible matrices unique to group $g$. 

\begin{figure}[!tb]
\centering
\includegraphics[width=\columnwidth]{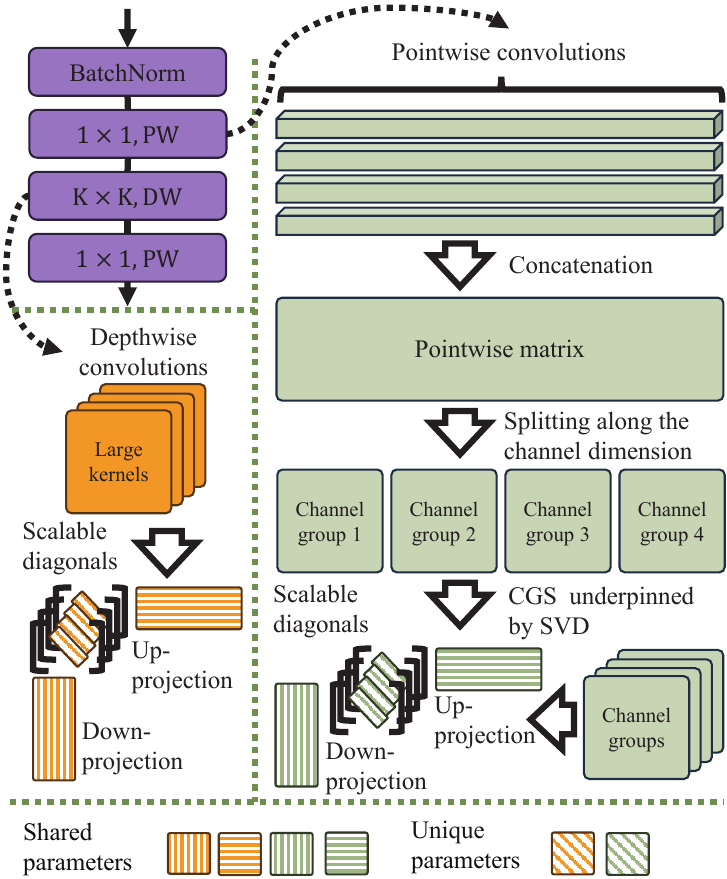}
\caption{Schematic of the CGS method. Pointwise convolution weights are partitioned into $k_{\text{auto}}$ submatrices, each approximated by shared low-rank projections and a group-specific diagonal. Depthwise convolution weights are approximated per channel using shared projections and a channel-specific diagonal.}
\label{method}
\end{figure}

\begin{figure}[t]
\centering
\includegraphics[width=1\columnwidth]{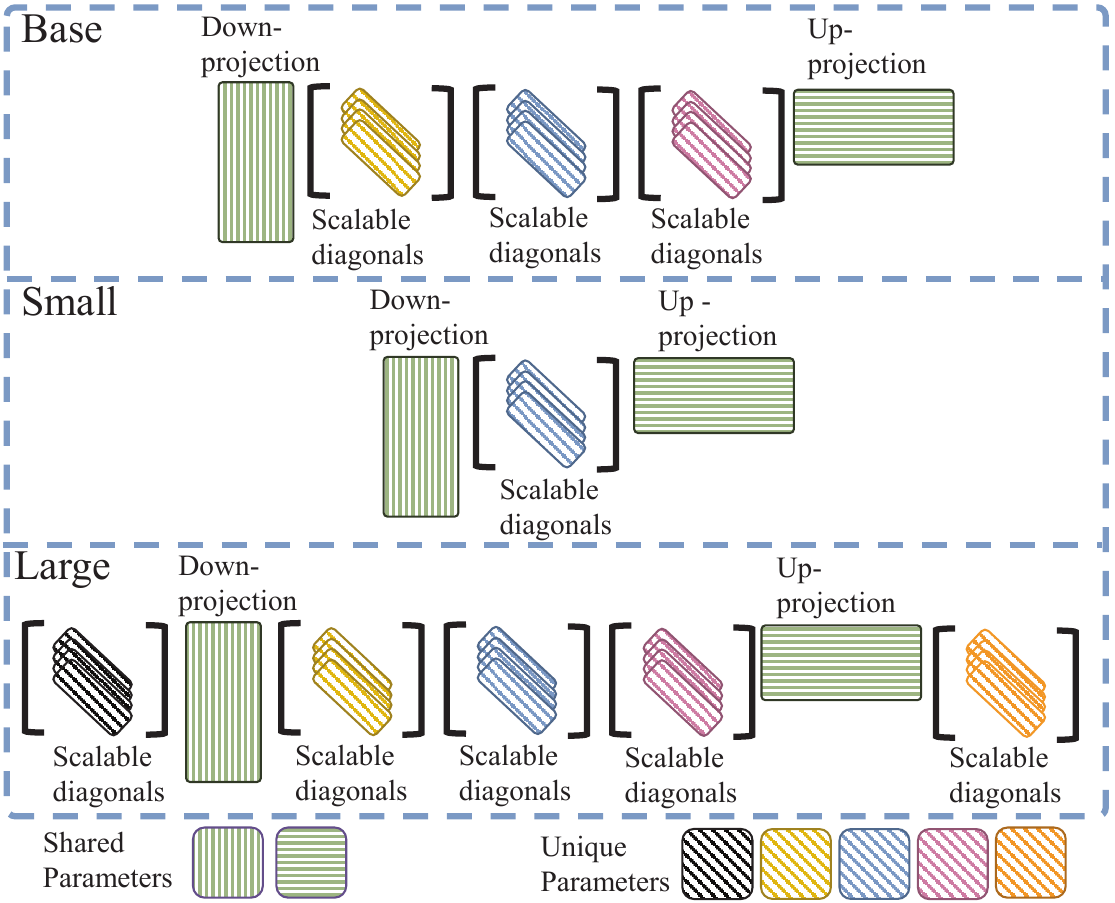}
\caption{Experimentally categorized parameter-sharing strategies across three complexity scales: (1) Base Configuration: assimilates outer invertible matrices into shared parameter space; (2) Small Configuration: employs complete matrix assimilation; (3) Large Configuration: directly implements the theoretically derived formulation with maximal parameter retention.}
\label{L/B/S_method}
\end{figure}

\subsection{Channel Group-Shared Low-Rank Approximation}

To implement this strategy in pointwise convolutions, we adopt a channel grouping scheme. An automatic group count $k_{\text{auto}}$ is designed based on the output-to-input channel ratio. This design removes the need for manual tuning while balancing model compression (reducing parameters) and representational capacity, establishing it as the baseline group number for subsequent experiments. Specifically, $k_{\text{auto}}$ is given by:

\begin{equation}
k_{\text{auto}} = \begin{cases} 
 C_{\text{in}} / C_{\text{out}}, & \text{if } C_{\text{in}} > C_{\text{out}} \\
C_{\text{out}} / C_{\text{in}}, & \text{otherwise}
\end{cases},
\label{eq:group_count}
\end{equation}
where $C_{\rm in}$ and $C_{\rm out}$ represent the number of input/output channels.

The pointwise convolution weight matrix $\mathbf{W}^{(P)} \in \mathbb{R}^{C_{\text{out}} \times C_{\text{in}}}$ is partitioned into $k_{\text{auto}}$ equal-sized group matrices along the input channel dimension:
\begin{equation}
\mathbf{W}^{(P)} = \left[ \mathbf{W}_1^{(P)} \mid \mathbf{W}_2^{(P)} \mid \cdots \mid \mathbf{W}_{k_{\text{auto}}}^{(P)} \right],
\end{equation}
where each group matrix $\mathbf{W}_i^{(P)} \in \mathbb{R}^{C_{\text{out}} \times C_{\text{out}}}$ for $i \in \{1, 2, \dots, k_{\text{auto}}\}$ when $C_{\text{in}} = k_{\text{auto}} \cdot C_{\text{out}}$. 

Furthermore, our method is also applicable to depthwise convolution. The depthwise convolution kernel $\mathbf{W}^{(D)} \in \mathbb{R}^{K_h \times K_w \times C}$, where $K_h$ and $K_w$ represent the height and width of the convolutional kernel respectively, is partitioned along the channel dimension into $C$ group kernels:
\begin{equation}
\mathbf{W}^{(D)} = \left[ \mathbf{W}_1^{(D)} \mid \mathbf{W}_2^{(D)} \mid \cdots \mid \mathbf{W}_C^{(D)} \right],
\end{equation}
where each group kernel $\mathbf{W}_i^{(D)} \in \mathbb{R}^{K_h \times K_w}$ for $i \in \{1, 2, \dots, C\}$. 

This identitical group partitioning scheme maintains architectural consistency with our pointwise convolution implementation. Unified dimensional factorization enables parameter sharing strategy across all channel groups, as shown in Fig.~\ref{method}.

Our derivation proceeds in two sequential steps:
\begin{enumerate}
    \item Application of Eq.~(\ref{Eq:svd}) to $\mathbf{W}_i^{(P)}$ and $\mathbf{W}_i^{(D)}$ yields:
    \begin{equation}
    \begin{split}
    \mathbf{W}_i^{(\cdot)} &= 
    \mathbf{U}^{(\cdot)}\mathbf{D}^{(\cdot)}\mathbf{V}^{\top(\cdot)},\\
    \end{split}
    \label{Eq:svd_app}
    \end{equation}
    where $\mathbf{W}_i^{(\cdot)}$ generically represents either $\mathbf{W}_i^{(P)}$ or $\mathbf{W}_i^{(D)}$ throughout the derivation.
    
    \item Subsequent application of Eq.~(\ref{Eq:orthogonal_transform}) to matrices $\mathbf{U}^{(\cdot)}$ and $\mathbf{V}^{\top(\cdot)}$ within Eq.~(\ref{Eq:svd_app}) produces:
    \begin{equation}
    \begin{split}
    \mathbf{W}_i^{(\cdot)} &= \mathbf{P}_{u}^{(\cdot)}\mathbf{W}^{(\cdot)}_{\rm shared\_u} \mathbf{Q}_{u}^{(\cdot)}\mathbf{D}^
    {(\cdot)}\mathbf{P}_{v}^{(\cdot)}\mathbf{W}^{(\cdot)}_{\rm shared\_v}\mathbf{Q}_{v}^{(\cdot)}.\\
\end{split}
    \label{Eq:final_derivation}
    \end{equation}
\end{enumerate}

To reduce the number of parameters, we let $\mathbf{P}$ and $\mathbf{Q}$ be diagonal matrices $\mathbf{D}_P$ and $\mathbf{D}_Q$, respectively. Eq.~(\ref{Eq:final_derivation}) can be rewritten as:
\begin{equation}
\begin{split}
    \mathbf{W}_i^{(\cdot)} &= \mathbf{D}_{P_u}^{(\cdot)}\mathbf{W}^{(\cdot)}_{\rm shared\_u} \mathbf{D}_{Q_u}^{(\cdot)}\mathbf{D}^
    {(\cdot)}\mathbf{D}_{P_v}^{(\cdot)}\mathbf{W}^{(\cdot)}_{\rm shared\_v} \mathbf{D}_{Q_v}^{(\cdot)}.
\end{split}
\label{Eq:represent_DWDDDWD}
\end{equation}
As illustrated in Fig.~\ref{L/B/S_method} (Large), the CGS-L method corresponds to the structure represented by Eq.~(\ref{Eq:represent_DWDDDWD}). Since $\mathbf{W}_{\rm shared}^{(\cdot)}$ are trainable parameters, to strike a balance between parameter count and model performance, we consider merging the outermost $\mathbf{D}_{P_u}^{(\cdot)}$ and $\mathbf{D}_{Q_v}^{(\cdot)}$ into $\mathbf{W}^{(\cdot)}_{\rm shared}$:
\begin{equation}
\begin{split}
    \mathbf{W}_i^{(\cdot)} &= \mathbf{W}^{(\cdot)'}_{\rm shared\_u}\mathbf{D}_{Q_u}^{(\cdot)}\mathbf{D}^
    {(\cdot)}\mathbf{D}_{P_v}^{(\cdot)}\mathbf{W}^{(\cdot)'}_{\rm shared\_v}.
\end{split}
\label{Eq:represent_WDDDW}
\end{equation}
As illustrated in Fig.~\ref{L/B/S_method} (Base), the CGS-B method corresponds to the structure represented by Eq.~(\ref{Eq:represent_WDDDW}). To further explore the flexibility of the approach and achieve a better trade-off between model parameter size and performance, we consider merging $\mathbf{D}_{Q_u}^{(\cdot)}$ and $\mathbf{D}_{P_v}^{(\cdot)}$ into $\mathbf{W}^{(\cdot)'}_{\rm shared}$:
\begin{equation}
\begin{split}
    \mathbf{W}_i^{(\cdot)} &= \mathbf{W}^{(\cdot)''}_{\rm shared\_u}\mathbf{D}^
    {(\cdot)}\mathbf{W}^{(\cdot)''}_{\rm shared\_v}.
\end{split}
\label{Eq:represent_WDW}
\end{equation}
As illustrated in Fig.~\ref{L/B/S_method} (Small), the CGS-S method corresponds to the structure represented by Eq.~(\ref{Eq:represent_WDW}). 

\begin{figure}[!tb]
    \centering
    \begin{subfigure}[b]{0.22\textwidth}
        \includegraphics[width=\textwidth]{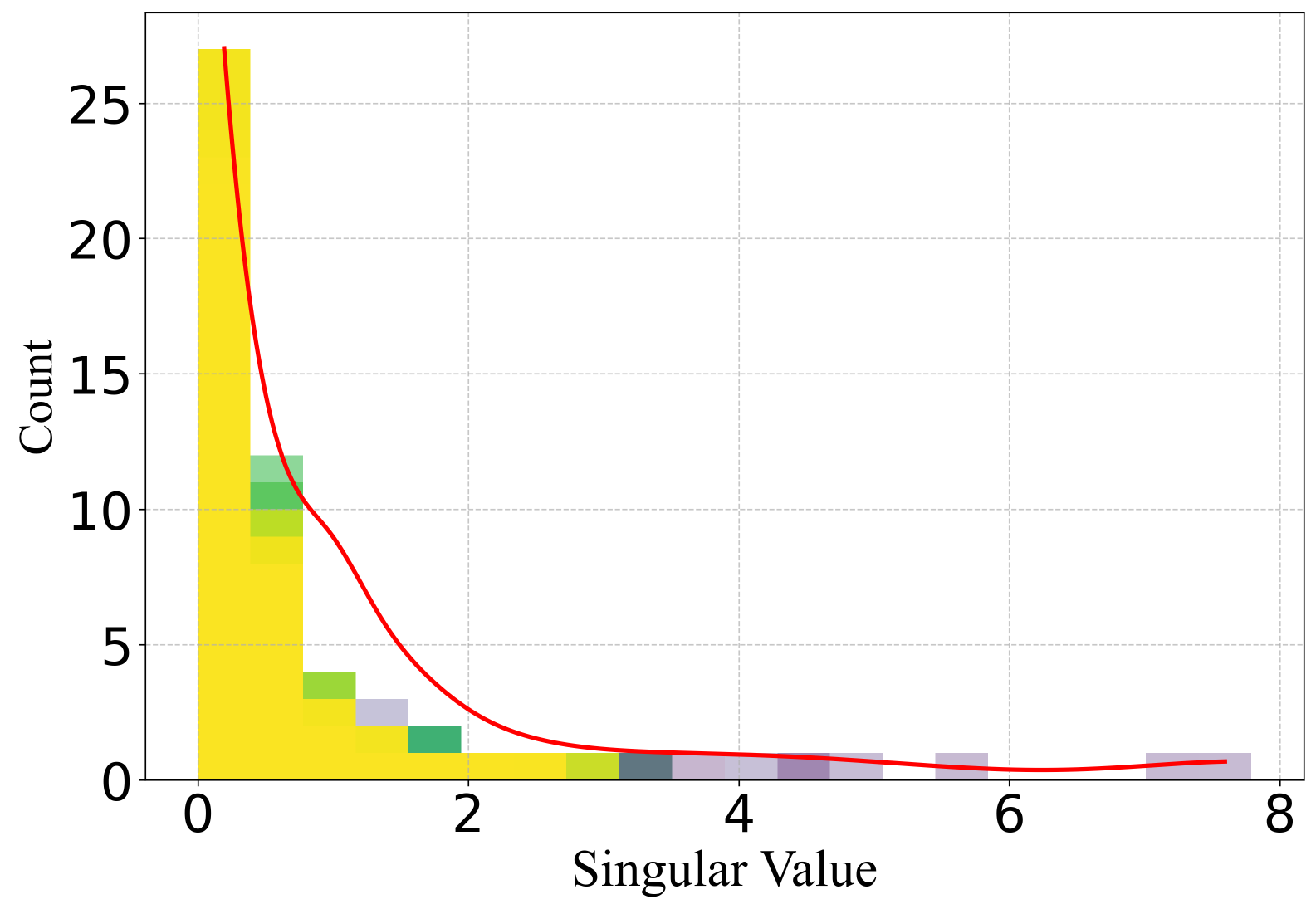}
        \caption{Depthwise convolution in Stage 3.}
        \label{fig:s3dw}
    \end{subfigure}
    \begin{subfigure}[b]{0.22\textwidth}
        \includegraphics[width=\textwidth]{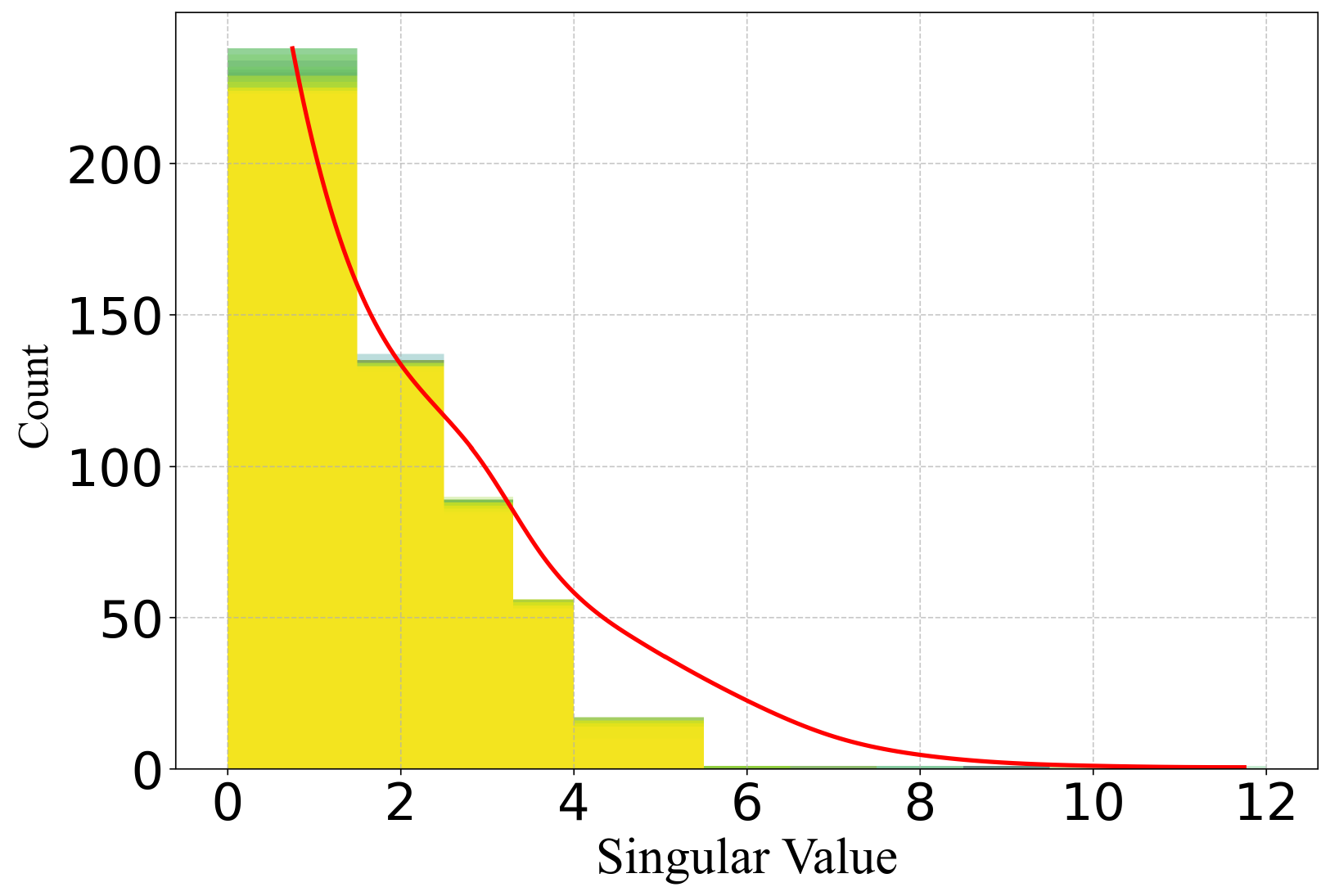}
        \caption{Pointwise convolution in Stage 3.}
        \label{fig:s3pw}
    \end{subfigure}
    \begin{subfigure}[b]{0.22\textwidth}
        \includegraphics[width=\textwidth]{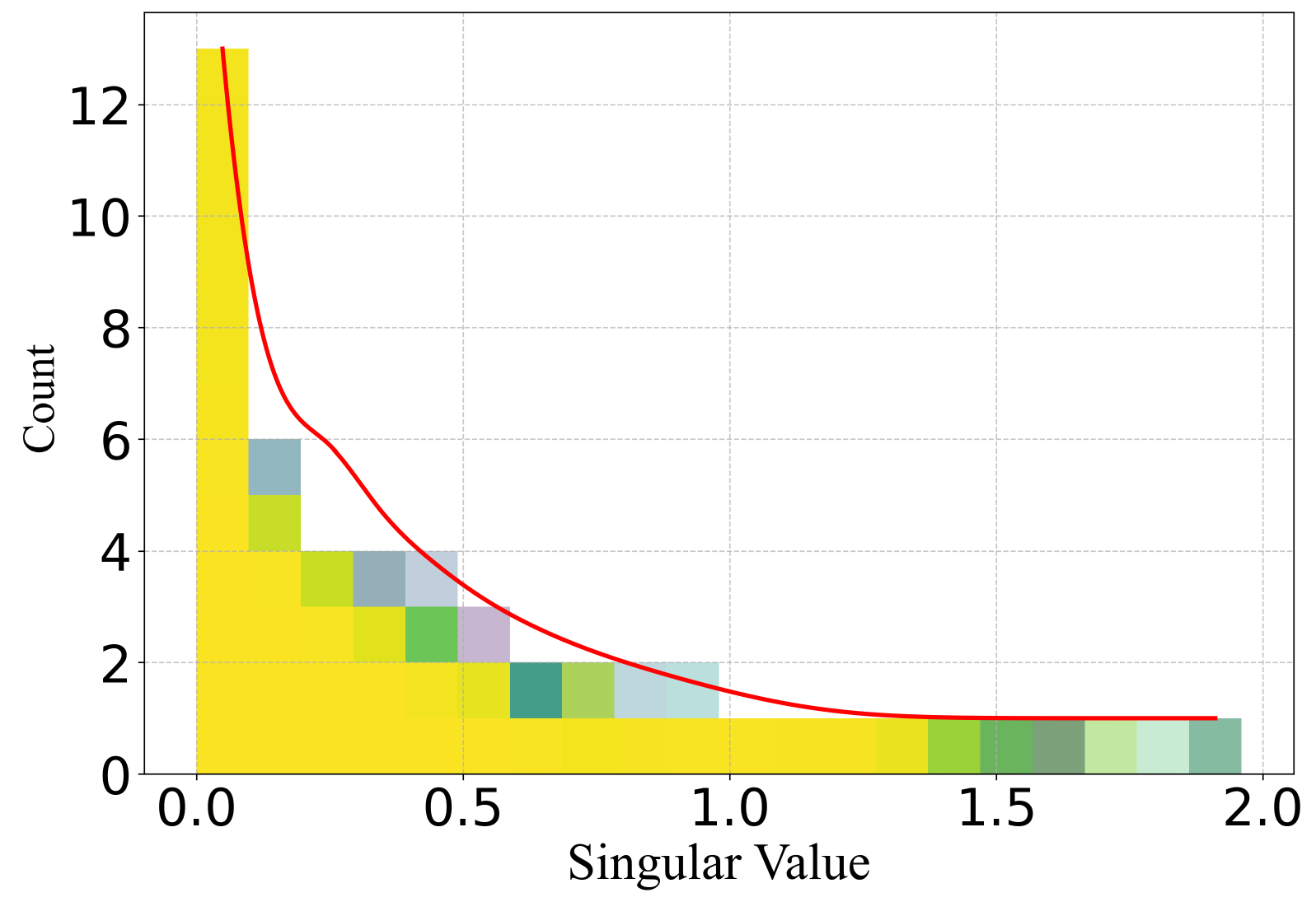}
        \caption{Depthwise convolution in Stage 4.}
        \label{fig:s4dw}
    \end{subfigure}
    \begin{subfigure}[b]{0.22\textwidth}
        \includegraphics[width=\textwidth]{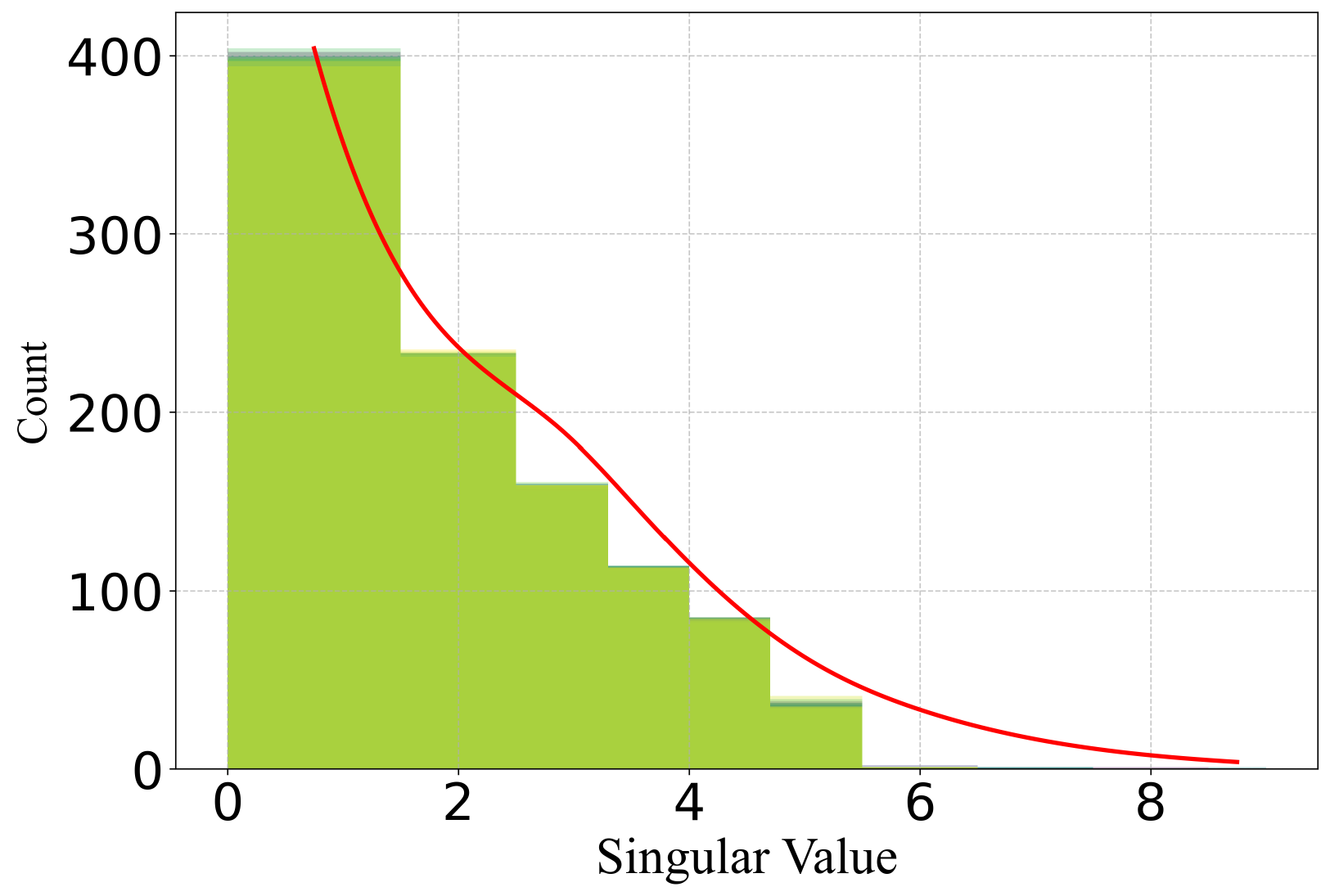}
        \caption{Pointwise convolution in Stage 4.}
        \label{fig:s4pw}
    \end{subfigure}
    \caption{Singular value distributions of pointwise and depthwise convolutional kernels in pre-trained RepLKNet-31B (Stages 3-4) after channel grouping. Each subfigure aggregates the kernel matrices grouped by stage and convolution type, with individual matrices color-coded. X-axis: Singular values; Y-axis: Count of singular values within defined bounds.}
    \label{fig:low-rank}
\end{figure}

The CGS compression strategy is applied selectively to Stages 3-4 of the model, as shown in Tab.~\ref{tab:Stage_results}, while earlier stages (1-2) are responsible for extracting low-level features and exhibit high sensitivity to compression. As shown in Fig.~\ref{fig:low-rank}, the analysis of the singular values from pointwise and depthwise convolutions in these stages of a pre-trained RepLKNet-31B model reveals a pronounced low-rank property and a power-law distribution, indicating that the weight matrices are inherently sparse and can be effectively reconstructed using a limited set of basis vectors. This observation directly motivates the design of the down/up-projection matrices as low-rank factorizations, achieving an effective balance between model performance and parameter reduction.

\subsection{Deployment on Mobile Devices}

\begin{figure}[tbph]
\centering
\includegraphics[width=\columnwidth]{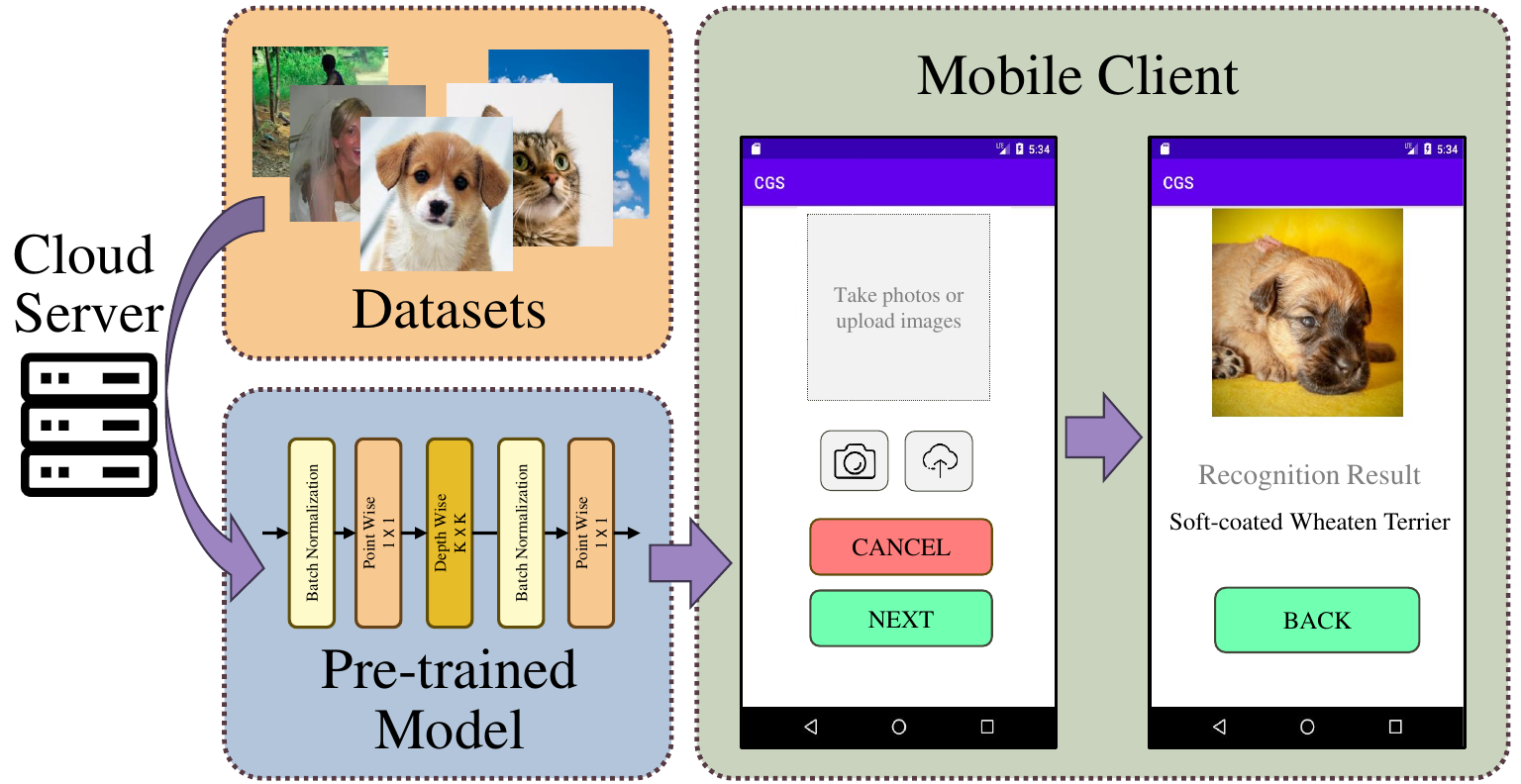}
\caption{This diagram schematically illustrates our edge deployment pipeline for pre-trained lightweight large-kernel convolutional foundation models: (1) pre-trained models optimized for mobile hardware are generated on cloud servers; (2) the compressed models are deployed to target mobile devices; (3) successful on-device inference execution is validated through real-world image inputs.}
\label{fig:bushu}
\end{figure}

This section outlines the deployment pipeline for pre-trained lightweight large-kernel convolutional foundation models on resource-constrained mobile devices. Models such as RepLKNet+CGS-B are first pre-trained on cloud servers using Graphics Processing Unit (GPU) acceleration. For deployment, the pre-trained weights are loaded with explicit Central Processing Unit (CPU) device mapping to ensure hardware compatibility, and the model architecture is programmatically reconstructed via parameter transplantation. The model is then converted into a static computation graph using TorchScript tracing with representative input tensors, and the hardware-agnostic binary is serialized for mobile execution. Finally, the compiled \texttt{.pt} file is integrated into an Android application via Android Studio~\cite{hagos2018android}. As empirically validated in Fig.~\ref{fig:bushu}, the deployed model performs accurate image classification on mobile devices, with real-time measurements confirming the practical benefits of CGS for edge deployment.
\section{Experiments}

We conduct extensive experiments to validate the effectiveness of the proposed lightweight sharing strategy. The evaluation encompasses multiple core vision tasks to assess generalization ability. In addition, we perform ablation studies to verify our design choices and investigate the impact of key hyperparameters. Finally, we conduct on-device experiments on mobile platforms to demonstrate practical feasibility.

\subsection{Experiments Setting}

\subsubsection{Datasets and Tasks}

We follow established evaluation protocols~\cite{liu2022convnet,liu2022more,ding2022scaling,chen2024pelk} across diverse tasks: large-scale classification (ImageNet-1K~\cite{deng2009imagenet}), fine-grained classification (CIFAR-100~\cite{krizhevsky2009learning}), scene parsing (ADE20K~\cite{zhou2019semantic}), and object-level recognition (COCO~\cite{lin2014microsoft}). This multi-task benchmark ensures a comprehensive assessment of model robustness and generalization.

\subsubsection{Baselines}

We compare against three state-of-the-art large-kernel CNN baselines: RepLKNet~\cite{ding2022scaling}, SLaK~\cite{liu2022more}, and ConvNeXt~\cite{liu2022convnet}. PeLK~\cite{chen2024pelk} is excluded due to the unavailability of its official implementation.

\subsubsection{CGS Variants}

To balance performance and efficiency, we instantiate three CGS variants of increasing capacity: CGS-S, CGS-B, and CGS-L. As illustrated in Fig.~\ref{L/B/S_method}, they correspond to the formulations in Eq.~\ref{Eq:represent_DWDDDWD}, Eq.~\ref{Eq:represent_WDDDW}, and Eq.~\ref{Eq:represent_WDW}, respectively. Enhanced representational capacity in the CGS-B and CGS-L configurations is achieved by incorporating additional stretching vectors into the down/up-projection matrices.

\subsubsection{Implementation Details}

Fig.~\ref{fig:pw_stage} shows the parameter distribution across stages for pre-trained RepLKNet-31B, ConvNeXt-B, and SLaK-B. A key observation is that parameters are heavily concentrated in the later stages (Stages 3-4), accounting for approximately 97\% of the total. As results in Tab.~\ref{tab:Stage_results} confirm, the early stages (1-2) are crucial for low-level feature extraction and highly sensitive to compression. Therefore, we apply the CGS compression strategy selectively only to Stages 3 and 4. All models are implemented in PyTorch and trained on NVIDIA A800 GPUs; detailed configurations are provided in Appendix~\ref{Training_Configurations}.

\begin{table}[!tb]
\centering
\caption{ImageNet-1K classification performance. Top-1 accuracy and parameter counts are compared across variants, demonstrating competitive results under similar model complexity.}
\label{tab:imagenet_results_same}
\begin{tabular}{l|c|c}
    \toprule
    Model & \makecell{Params (M)} & \makecell{Top-1 Acc} \\
    \midrule
    ResNet18~\cite{he2016deep} & 11.7 & 69.8 \\
    PoolFormer-S12~\cite{yu2022metaformer} & 11.9 & 77.2 \\
    \textbf{RepLKNet+CGS-S (Ours)} & 14.1 & 79.1 \\
    \midrule
    PVT-Tiny~\cite{wang2021pyramid} & 13.2 & 75.1 \\
    PVTv2-B1~\cite{wang2022pvt} & 13.1 & 78.7 \\
    \textbf{RepLKNet+CGS-B (Ours)} & 14.7 & 80.0 \\
    \midrule
    RegNetY-4G~\cite{radosavovic2020designing} & 21.0 & 80.0 \\
    DeiT-Small/16~\cite{touvron2021training} & 22.1 & 79.8 \\
    T2T-ViT$_t$-14~\cite{yuan2021tokens} & 21.5 & 81.7 \\
    gMLP-S~\cite{liu2021pay} & 20.0 & 79.6 \\
    PoolFormer-S24~\cite{yu2022metaformer} & 21.4 & 80.3 \\
    \textbf{RepLKNet+CGS-L (Ours)} & 17.1 & 82.1 \\
    \bottomrule
\end{tabular}
\vspace{-0.05cm}
\end{table}

\begin{table}[!tb]
\centering
\caption{ImageNet-1K classification performance. Top-1 accuracy and parameter counts are compared across variants, with our method achieving an empirically favorable accuracy-parameter trade-off.}
\label{tab:imagenet_results}
\begin{tabular}{l|c|c}
    \toprule
    Model & \makecell{Params (M)} & \makecell{Top-1 Acc} \\
    \midrule
    T2T-ViT$_{t}$-14~\cite{yuan2021tokens} & 22 & 81.7 \\
    PerViT-S~\cite{min2022peripheral} & 21 & 82.1 \\
    Swin-T~\cite{liu2021swin} & 28 & 81.3 \\
    ConvNeXt-T~\cite{liu2022convnet} & 29 & 82.1 \\
    SLaK-T~\cite{liu2022more} & 30 & 82.5 \\
    PeLK-T~\cite{chen2024pelk} & 29 & 82.6 \\
    \textbf{RepLkNet+CGS-S (Ours)} & 14 & 79.1 \\
    \midrule
    PVT-Large~\cite{wang2021pyramid} & 61 & 81.7 \\
    T2T-ViT$_{t}$-19~\cite{yuan2021tokens} & 39 & 82.4 \\
    PerViT-M~\cite{min2022peripheral} & 44 & 82.9 \\
    Swin-S~\cite{liu2021swin} & 50 & 83.0 \\
    ConvNeXt-S~\cite{liu2022convnet} & 50 & 83.1 \\
    SLaK-S~\cite{liu2022more} & 55 & 83.8 \\
    PeLK-S~\cite{chen2024pelk} & 50 & 83.9 \\
    \textbf{RepLKNet+CGS-B (Ours)} & 15 & 80.0 \\
    \midrule
    DeiT-B/16~\cite{touvron2021training} & 87 & 81.8 \\
    RepLKNet-31B~\cite{ding2022scaling} & 79 & 83.5 \\
    Swin-B~\cite{liu2021swin} & 88 & 83.5 \\
    ConvNeXt-B~\cite{liu2022convnet} & 89 & 83.8 \\
    SLaK-B~\cite{liu2022more} & 95 & 84.0 \\
    PeLK-B~\cite{chen2024pelk} & 89 & 84.2 \\
    \textbf{RepLKNet+CGS-L (Ours)} & 17 & 82.1 \\
    \bottomrule
\end{tabular}
\end{table}

\subsection{Experimental Comparisons}

\subsubsection{ImageNet Classification}

We evaluate CGS-S/B/L integrated into RepLKNet on ImageNet-1K~\cite{deng2009imagenet}. As shown in Tab.~\ref{tab:imagenet_results_same}, all variants achieve better performance under comparable parameter budgets, with CGS-L attaining superior accuracy at lower cost. A broader comparison against SLaK~\cite{liu2022more} and ConvNeXt~\cite{liu2022convnet} (Tab.~\ref{tab:imagenet_results}) confirms that CGS significantly reduces parameters while maintaining competitive accuracy.

Overall, CGS-B achieves an favorable efficiency-accuracy trade-off. Compared to PVT-Tiny~\cite{wang2021pyramid} at a similar scale, it improves Top-1 accuracy by 6.5\% with only a 1.5M parameters increase. Against RepLKNet-31B~\cite{ding2022scaling}, it reduces parameters by 81\% while limiting the accuracy drop to 4.2\%. These results validate that our lightweight sharing strategy effectively compresses models without compromising performance.

\subsubsection{Object Detection}

We evaluate object detection performance on COCO~\cite{lin2014microsoft} using Cascade Mask R-CNN~\cite{cai2018cascade, he2017mask, sfm_tpami2018} with a RepLKNet backbone. Following standard practice in MMDetection~\cite{chen2019mmdetection} and ConvNeXt~\cite{liu2022convnet}, we use the AdamW optimizer with multi-scale training over 36 epochs.

Integrating the CGS-B strategy yields consistent gains under comparable parameter budgets (Tab.~\ref{tab:COCO_results_1}). In comparisons against SLaK~\cite{liu2022more} and ConvNeXt~\cite{liu2022convnet} (Tab.~\ref{tab:COCO_results_2}), CGS achieves competitive detection accuracy while reducing parameters substantially, with a reduction of over 47\% for RepLKNet-31B. These results demonstrate that our lightweight sharing mechanism maintains strong detection capability while significantly improving model efficiency.

\begin{table}[!tb]
\centering
\caption{MS COCO object detection performance. The CGS-B approach maintains competitive accuracy at comparable model complexity.}
\label{tab:COCO_results_1}
\begin{tabular}{l|c|c|c}
    \toprule 
    Method & \makecell{Params \\ (M)} & $\text{AP}^\text{box}$ & $\text{AP}^\text{mask}$ \\
    \midrule
    ResNet101~\cite{he2016deep} & 63 & 40.4 & 36.4 \\
    ResNeXt101-32x4d~\cite{xie2017aggregated} & 63 & 41.9 & 37.5 \\
    PVT-Medium~\cite{wang2021pyramid} & 64 & 42.0 & 39.0 \\
    Swin-T~\cite{liu2021swin} & 86 & 50.5 & 43.7 \\
    ConvNeXt-T~\cite{liu2022convnet} & 86 & 50.4 & 43.7 \\
    PeLK-T~\cite{chen2024pelk} & 86 & 51.4 & 44.6 \\
    \textbf{RepLKNet+CGS-B (Ours)} & 72 & 49.8 & 43.5 \\
    \bottomrule
\end{tabular}
\end{table}

\begin{table}[!tb]
\centering
\caption{MS COCO object detection performance. CGS-B strikes a competitive balance between efficiency and accuracy among the evaluated methods.}
\label{tab:COCO_results_2}
\begin{tabular}{l|c|c|c}
    \toprule
    Method & \makecell{Params \\ (M)} & $\text{AP}^\text{box}$ & $\text{AP}^\text{mask}$ \\
    \midrule
    Swin-S~\cite{liu2021swin} & 107 & 51.8 & 44.7 \\
    ConvNeXt-S~\cite{liu2022convnet} & 108 & 51.9 & 45.0 \\
    PeLK-S~\cite{chen2024pelk} & 108 & 52.2 & 45.3 \\
    Swin-B~\cite{liu2021swin} & 145 & 51.9 & 45.0 \\
    RepLKNet-31B~\cite{ding2022scaling} & 137 & 52.2 & 45.2 \\
    SLaK-B~\cite{liu2022more} & 152 & 52.5 & 45.5 \\
    ConvNeXt-B~\cite{liu2022convnet} & 146 & 52.7 & 45.6 \\
    PeLK-B~\cite{chen2024pelk} & 147 & 52.9 & 45.9 \\
    \textbf{RepLKNet+CGS-B (Ours)} & 72 & 49.8 & 43.5 \\
    \bottomrule
\end{tabular}
\end{table}

\subsubsection{Semantic Segmentation}

We evaluate semantic segmentation performance on the ADE20K dataset using UperNet~\cite{xiao2018unified} with a RepLKNet backbone pre-trained on ImageNet-1K and fine-tuned for 160K iterations. As shown in Tab.~\ref{tab:ADE20K_results}, our parameter-sharing strategy strikes an empirically favorable balance: it maintains competitive single-scale mean Intersection-over-Union (mIoU) while reducing parameters by 57\% compared to the original RepLKNet-31B. This demonstrates that the approach preserves the large-kernel advantage in global receptive fields while significantly improving efficiency for dense prediction tasks.

\begin{table}[!tb]
\centering
\caption{ADE20K semantic segmentation performance. CGS-B achieves an empirically favorable accuracy-parameter efficiency trade-off.}
\label{tab:ADE20K_results}
\begin{tabular}{l|c|c}
    \toprule
    Method & \makecell{Params (M)} & \makecell{mIoU (\%)} \\
    \midrule
    Swin-T~\cite{liu2021swin} & 60 & 44.5 \\
    ConvNeXt-T~\cite{liu2022convnet} & 60 & 46.0 \\
    SLaK-T~\cite{liu2022more} & 64 & 47.6 \\
    PeLK-T~\cite{chen2024pelk} & 62 & 48.1 \\ 
    Swin-S~\cite{liu2021swin} & 81 & 47.6 \\
    ConvNeXt-S~\cite{liu2022convnet} & 82 & 48.7 \\
    SLaK-S~\cite{liu2022more} & 89 & 49.4 \\
    PeLK-S~\cite{chen2024pelk} & 84 & 49.7 \\
    Swin-B~\cite{liu2021swin} & 121 & 48.1 \\
    ConvNeXt-B~\cite{liu2022convnet} & 122 & 49.1 \\
    RepLKNet-31B~\cite{ding2022scaling} & 112 & 49.9 \\
    SLaK-B~\cite{liu2022more} & 131 & 50.2 \\
    PeLK-B~\cite{chen2024pelk} & 126 & 50.4 \\
    \textbf{RepLKNet+CGS-B (Ours)} & 48 & 45.3 \\
    \bottomrule
\end{tabular}
\end{table}

\subsection{Ablation Studies}

This section conducts systematic ablation studies to validate our design choices and examine the impact of key hyperparameters.

\subsubsection{Rationale for Stage-Specific Compression}

Restricting CGS compression to Stages 3-4 is a deliberate design choice. As shown in Fig.~\ref{fig:pw_stage}, these stages contain approximately 97\% of the model parameters. In contrast, Stages 1-2 extract low-level features and are highly sensitive to compression. Ablation results in Tab.~\ref{tab:Stage_results} confirm that, compared to compressing only Stages 3-4 which preserves accuracy, applying CGS-B to all stages of RepLKNet leads to a relative accuracy reduction of 1.6\% while saving merely 1.3M parameters, indicating that concentrating compression on the parameter-dense later stages yields an empirically favorable trade-off between efficiency and accuracy.

\begin{table}[!tb]
\centering
\caption{Comparative evaluation of classification performance on ImageNet-1K using CGS across different stage ranges.}
\label{tab:Stage_results}
\begin{tabular}{l|c|c|c}
    \toprule
    Model & Satge & \makecell{Params (M)} & \makecell{Top-1 Acc} \\
    \midrule
    \multirow{2}{*}{\textbf{RepLKNet+CGS-B}} & 3-4 & 14.7 & 80.0 \\
                                             & 1-4 & 13.4 & 78.7 \\
    \bottomrule
\end{tabular}
\end{table}

\subsubsection{Analysis of Bottleneck Dimension and Group Count \texorpdfstring{$k$}{k}}

\begin{figure}[t]
\centering
\includegraphics[width=\columnwidth]{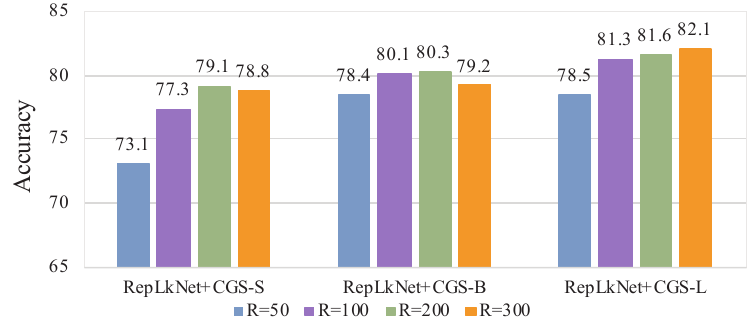}
\caption{Systematic evaluation of dimensionality scaling effects in down/up-projection matrices on CGS performance for CIFAR-100 datasets. Detailed results are presented in Appendix~\ref{app:Ablation_study_dim}.}
\label{fig:dimension}
\end{figure}

We conduct a systematic study on CIFAR-100 to evaluate the impact of two key hyperparameters in CGS: the bottleneck dimension $R$ of the down/up-projection matrices and the group count $k$.

Using RepLKNet as the backbone, we test bottleneck dimensions \(R \in \{50, 100, 200, 300\}\) across the CGS-S, -B, and -L variants. Fig.~\ref{fig:dimension} shows that \(R=200\) achieves an empirically favorable balance between model complexity and accuracy for all variants. Subsequently, with \(R\) fixed at 200, we ablate the group count \(k\) relative to the automatically determined \(k_{\text{auto}}\). As shown in Tab.~\ref{tab:Ablation_results_k}, performance degrades when \(k\) deviates from \(k_{\text{auto}}\). A smaller \(k\) limits the adaptation flexibility because each shared transformation must model a larger portion of the pointwise weight matrix. In contrast, a large \(k\) results in many fine-grained submatrices, reducing the structural information contained within each partition. Consequently, sharing the same transformation across these fragmented submatrices becomes less effective, limiting the quality of the learned weight updates and ultimately degrading accuracy. Therefore, \(k_{\text{auto}}\) provides a robust empirical trade-off between adaptation flexibility and parameter sharing.

\subsubsection{Cross-Architecture Generalization} 

To evaluate the generality of CGS within its target application domain, we deploy the CGS-B variant on three representative large-kernel CNN backbones: RepLKNet~\cite{ding2022scaling}, ConvNeXt~\cite{liu2022convnet}, and SLaK~\cite{liu2022more}. Tab.~\ref{tab:imagenet_results_cross} shows that CGS-B achieves an empirically favorable accuracy-parameter trade-off, compressing parameters by $>$77\% with accuracy drops below 6.5\% across all evaluated architectures. These results demonstrate that CGS generalizes well across diverse large-kernel CNN architectures, supporting its effectiveness within the target deployment setting.

\begin{table}[t]
\centering
\caption{Systematic evaluation of Group Count ($k$) in down/up-projection matrices on CGS performance for CIFAR-100 datasets.}
\begin{tabular}{l|c|c|c}
    \toprule
    Model & $k$ relative & \makecell{Params \\ (M)} & \makecell{Top-1 \\ Acc} \\
    \midrule
    \multirow{3}{*}{RepLkNet+CGS-S} & $0.5k_{\text{auto}}$ & 17.8 & 78.2 \\
                                     & $k_{\text{auto}}$ & 13.2 & 79.1 \\
                                     & $2k_{\text{auto}}$ & 11.0 & 76.1 \\
    \midrule
    \multirow{3}{*}{RepLkNet+CGS-B} & $0.5k_{\text{auto}}$ & 18.2 & 78.2 \\
                                     & $k_{\text{auto}}$ & 13.7 & 80.3 \\
                                     & $2k_{\text{auto}}$ & 11.6 & 76.4 \\
    \midrule
    \multirow{3}{*}{RepLkNet+CGS-L} & $0.5k_{\text{auto}}$ & 20.6 & 78.9 \\
                                     & $k_{\text{auto}}$ & 16.2 & 81.6 \\
                                     & $2k_{\text{auto}}$ & 14.2 & 76.3 \\
    \bottomrule
\end{tabular}
\label{tab:Ablation_results_k}
\end{table}

\begin{table}[!tb]
\centering
\caption{ImageNet-1K classification performance. RepLKNet, ConvNeXt, and SLaK backbones integrated with the CGS-B compression.}
\label{tab:imagenet_results_cross}
\begin{tabular}{l|c|c}
    \toprule
    Model & \makecell{Params (M)} & \makecell{Top-1 Acc} \\
    \midrule
    RepLKNet-31B~\cite{ding2022scaling} & 79 & 83.5 \\
    \textbf{RepLKNet+CGS-B (Ours)} & 15 & 80.0 \\
    \midrule
    ConvNeXt-B~\cite{liu2022convnet} & 89 & 83.8 \\
    \textbf{ConvNeXt+CGS-B (Ours)} & 20 & 78.5 \\
    \midrule
    SLaK-B~\cite{liu2022more} & 95 & 84.0 \\
    \textbf{SLaK+CGS-B (Ours)} & 17 & 79.8 \\
    \bottomrule
\end{tabular}
\end{table}

\subsubsection{Ablation on Structural and Quantization Synergy}

\begin{figure}[t]
    \centering
    \begin{subfigure}[t]{0.495\linewidth}
        \centering
        \includegraphics[width=\linewidth]{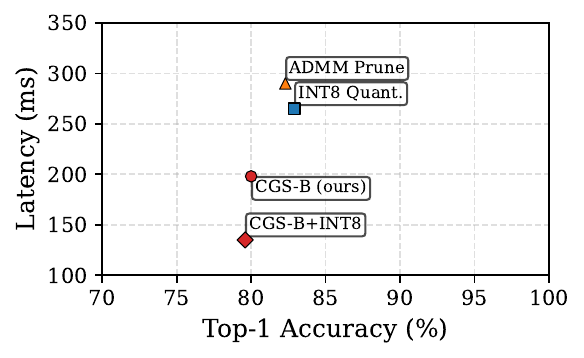}
        \caption{Accuracy vs.\ Latency}
        \label{fig:deploy_lat}
    \end{subfigure}
    \hfill
    \begin{subfigure}[t]{0.495\linewidth}
        \centering
        \includegraphics[width=\linewidth]{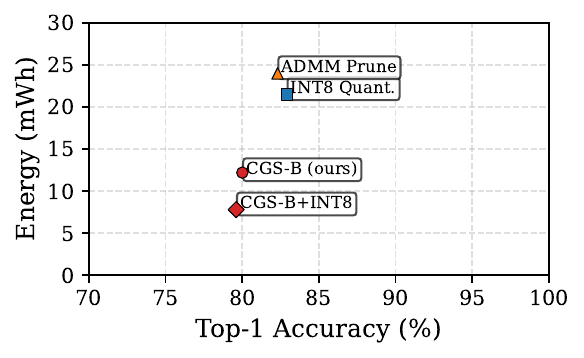}
        \caption{Accuracy vs.\ Energy}
        \label{fig:deploy_energy}
    \end{subfigure}
    \caption{Performance comparison on a Snapdragon 8 Gen 3 mobile platform.}
    \label{fig:deploy_tradeoff}
\end{figure}

Fig.~\ref{fig:deploy_tradeoff} compares on-device efficiency. CGS-B cuts latency by 25\% and energy by 43\% over the INT8 baseline, with a moderate accuracy drop. Combined with INT8, CGS-B+INT8 achieves 135\,ms latency and 7.8\,mWh energy while incurring only a 0.5\% relative accuracy drop, showing strong structural-quantization synergy. In contrast, ADMM pruning, despite fewer parameters, yields higher latency and energy due to irregular sparsity. Detailed results are in Appendix~\ref{app:deploy_struc_quant}.

\begin{table}[t]
\centering
\caption{Comparison of deployment performance of different large-kernel convolution models on various mobile devices, including RAM and loading time.}
\label{tab:different_devices_results}
    \begin{tabular}{l|c|c|c}
    \toprule
    Device & Model & \makecell{RAM \\ (M)} & \makecell{Loading \\ Time (ms)}\\
    \midrule
    \multirow{4}{*}{\shortstack[l]{Google \\ Pixel 2 \\ (Emulator)}} & ConvNeXt-B~\cite{liu2022convnet} & 372.17 & 1541 \\
                                                    & RepLKNet-31B~\cite{ding2022scaling} & 344.75 & 1313 \\
                                                    & SLaK-B~\cite{liu2022more} & 641.08 & 3720 \\
                                                    & \textbf{RepLKNet+CGS-B} & 96.58 & 591 \\
    \midrule
    \multirow{4}{*}{\shortstack[l]{Galaxy \\ Nexus \\ (Emulator)}} & ConvNeXt-B~\cite{liu2022convnet} & 372.67 & 2518 \\
                                                  & RepLKNet-31B~\cite{ding2022scaling} & 345.63 & 2639 \\
                                                  & SLaK-B~\cite{liu2022more} & 633.42 & 4443 \\
                                                  & \textbf{RepLKNet+CGS-B} & 96.95 & 609 \\
    \midrule
    \multirow{4}{*}{\shortstack[l]{HUAWEI \\ P20 pro}} & ConvNeXt-B~\cite{liu2022convnet} & 385.84 & 2854 \\
                                                    & RepLKNet-31B~\cite{ding2022scaling} & 360.17 & 3383 \\
                                                    & SLaK-B~\cite{liu2022more} & 666.80 & 5167 \\
                                                    & \textbf{RepLKNet+CGS-B} & 112.68 & 1482 \\
    \midrule
    \multirow{4}{*}{\shortstack[l]{HUAWEI \\ nova 8}} & ConvNeXt-B~\cite{liu2022convnet} & 399.63 & 1725 \\
                                                  & RepLKNet-31B~\cite{ding2022scaling} & 370.16 & 1912 \\
                                                  & SLaK-B~\cite{liu2022more} & 666.96 &  3019 \\
                                                  & \textbf{RepLKNet+CGS-B} & 125.14 & 918 \\
    \midrule
    \multirow{4}{*}{\shortstack[l]{Honor \\ 200 Pro}} & ConvNeXt-B~\cite{liu2022convnet} & 399.67 & 1761 \\
                                                  & RepLKNet-31B~\cite{ding2022scaling} & 377.50 & 1613 \\
                                                  & SLaK-B~\cite{liu2022more} & 673.13 &  3012 \\
                                                  & \textbf{RepLKNet+CGS-B} & 120.18 & 624 \\
    \midrule
    \multirow{4}{*}{\shortstack[l]{OPPO \\ Reno4}} & ConvNeXt-B~\cite{liu2022convnet} & 394.02 & 1855 \\
                                                  & RepLKNet-31B~\cite{ding2022scaling} & 372.26 & 2080 \\
                                                  & SLaK-B~\cite{liu2022more} & 668.22 &  3197 \\
                                                  & \textbf{RepLKNet+CGS-B} & 125.82 & 1004 \\
    \bottomrule
\end{tabular}
\end{table}

\begin{figure}[tbph]
\centering
\includegraphics[width=\columnwidth]{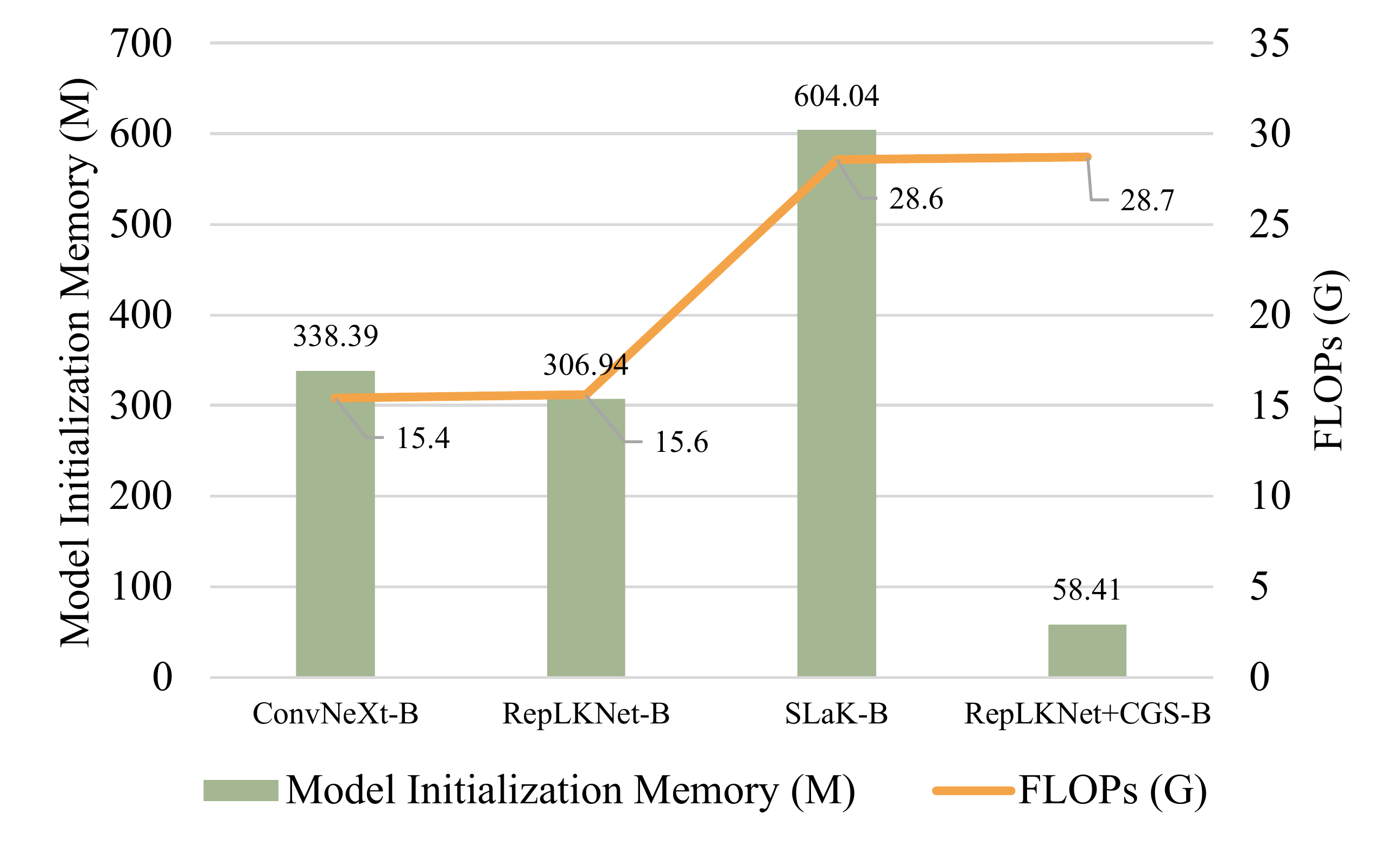}
\caption{Comparison of memory footprint during model initialization and total FLOPs for inference across different models on mobile devices.}
\label{fig:flops_M}
\end{figure}

\subsection{Mobile Devices Performance}

This section evaluates the deployment of pre-trained large-kernel foundation models on resource-constrained mobile devices, assessing the practicality of our lightweight CGS approach. The specific devices and processors used are detailed in Appendix~\ref{Mobile}.

\subsubsection{Memory Efficiency Benchmark}

We integrate CGS into several large-kernel convolutional models and deploy them on various mobile devices. Tab.~\ref{tab:different_devices_results} and Fig.~\ref{fig:flops_M} report key metrics including Random Access Memory (RAM) usage, initialization memory, model loading time, and inference FLOPs. Our method significantly reduces model loading time compared to alternatives, while maintaining Floating-point Operations (FLOPs) only slightly higher than other methods. This demonstrates an effective trade-off that prioritizes storage and loading efficiency, which are often critical bottlenecks in mobile deployment, with low computational overhead.

\subsubsection{On-Device Latency and Energy Profiling}

We conduct device-level measurements on three representative Android devices with different mainstream processors. Results in Tab.~\ref{tab:real-world_mobile_result} show that CGS-B achieves lower inference latency despite higher FLOPs. Wireless evaluation results are reported as the average of multiple independent trials. This counter-intuitive result stems from the memory-bound nature of modern mobile Neural Processing Units (NPUs), where data movement, not arithmetic, is the primary bottleneck~\cite{chen2014diannao}. CGS addresses this via four key optimizations: (1) compressing pointwise convolutions to reduce model footprint and weight transfer; (2) sharing projection matrices to minimize random memory access and improve cache efficiency; (3) employing a low-rank representation to reduce activation bandwidth; and (4) restructuring operations into larger, more regular GEMM tasks for better NPU acceleration. The per-run latency variation remained below 5\%, confirming stable performance across hardware.

\begin{table}[!tb]
\centering
\caption{Mobile deployment performance. Latency, energy consumption, memory footprint, and FPS are reported for different models.}
\label{tab:real-world_mobile_result}
\setlength{\tabcolsep}{1pt}
\begin{tabular}{l|c|c|c|c|c}
    \toprule
    Device & Model & \makecell{Lat. \\ (ms)} & \makecell{Energy \\ (mWh)} & \makecell{Mem. \\ (MB)} & FPS \\
    \midrule
    \multirow{4}{*}{\makecell{OnePlus \\ 13}} 
        & ConvNeXt-T~\cite{liu2022convnet} & $215 \pm 5$ & $13.8 \pm 0.3$ & 141 & 4.7 \\
        & RepLKNet-B~\cite{ding2022scaling} & $242 \pm 6$ & $16.3 \pm 0.4$ & 338 & 4.1 \\
        & \textbf{\makecell{RepLKNet-B \\ + CGS-B~(Ours)}} & $198 \pm 5$ & $12.2 \pm 0.3$ & 115 & 5.0 \\
    \midrule
    \multirow{4}{*}{\makecell{Xiaomi \\ 14 Ultra}} 
        & ConvNeXt-T~\cite{liu2022convnet} & $208 \pm 4$ & $13.4 \pm 0.3$ & 145 & 4.8 \\
        & RepLKNet-B~\cite{ding2022scaling} & $235 \pm 5$ & $15.7 \pm 0.4$ & 345 & 4.3 \\
        & \textbf{\makecell{RepLKNet-B \\ + CGS-B~(Ours)}} & $189 \pm 4$ & $11.8 \pm 0.3$ & 118 & 5.3 \\
    \midrule
    \multirow{4}{*}{\makecell{vivo \\ X300 Pro}} 
        & ConvNeXt-T~\cite{liu2022convnet} & $205 \pm 4$ & $13.0 \pm 0.2$ & 150 & 5.1 \\
        & RepLKNet-B~\cite{ding2022scaling} & $228 \pm 5$ & $15.1 \pm 0.3$ & 392 & 4.4 \\
        & \textbf{\makecell{RepLKNet-B \\ + CGS-B~(Ours)}} & $186 \pm 4$ & $11.3 \pm 0.2$ & 121 & 5.4 \\
    \bottomrule
\end{tabular}
\end{table}

\subsubsection{Wireless Offloading Performance Analysis}

To evaluate CGS in edge-computing scenarios, we test end-to-end performance under a wireless ``device-to-cloud'' offloading setup, comparing 5G and Wi-Fi 6 networks, with ConvNeXt-T and RepLKNet-B as baselines. The total latency includes model load, deserialization, inference, and the associated communication overhead. As shown in Tab.~\ref{tab:wireless_results}, wireless communication consistently contributes 54\%--62\% of the overall latency across all evaluated models. This observation suggests that, in the considered deployment scenario, reducing model size has a greater impact on end-to-end latency than reducing FLOPs alone.

CGS compresses RepLKNet-B~\cite{ding2022scaling} from 306.94 MB to 58.41 MB, reducing transmitted data from 287 MB to 172 MB, and cuts end-to-end latency by ~27\% on 5G and ~25\% on Wi-Fi 6. It also outperforms lightweight ConvNeXt-T~\cite{liu2022convnet}, with 46.5\% smaller model size and 13-14\% lower latency. This confirms that CGS effectively mitigates the high-communication-latency bottleneck, making it suitable for efficient edge-to-cloud collaborative inference.

\begin{table}[!tb]
\centering
\caption{The evaluation of end-to-end wireless execution was conducted under two distinct network conditions: 5G connectivity was tested on a Xiaomi 14 Ultra device, while Wi-Fi 6 performance was measured on a OnePlus 13 device. The column ``Offload (ms)'' denotes on-device computation latency.}
\label{tab:wireless_results}
\setlength{\tabcolsep}{2pt}
\begin{tabular}{c|c|c|c|c|c}
    \toprule
    Network & Model & \makecell{Size \\ (MB)} & \makecell{Offload \\ (ms)} & \makecell{Comm. \\ (MB)} & \makecell{Total \\ (ms)} \\
    \midrule
    \multirow{4}{*}{5G} 
        & ConvNeXt-T~\cite{liu2022convnet} & 109.30 & 145 & 102 & 355 \\
        & RepLKNet-B~\cite{ding2022scaling} & 306.94 & 178 & 287 & 420 \\
        & \textbf{\makecell{RepLKNet-B \\ + CGS-B~(Ours)}} & 58.41 & 116 & 172 & 305 \\
    \midrule
    \multirow{4}{*}{Wi-Fi 6} 
        & ConvNeXt-T~\cite{liu2022convnet} & 109.30 & 170 & 102 & 380 \\
        & RepLKNet-B~\cite{ding2022scaling} & 306.94 & 203 & 287 & 438 \\
        & \textbf{\makecell{RepLKNet-B \\ + CGS-B~(Ours)}} & 58.41 & 142 & 172 & 330 \\
    \bottomrule
\end{tabular}
\end{table}
\section{Related Works}

\subsection{Large-Kernel Convolutions}

The evolution of Convolutional Neural Networks (CNNs)~\cite{lecun1989backpropagation, bello2019attention, carion2020end, sfm_cvpr15, huang2019ccnet, lecun2002gradient, wang2022pvt} has witnessed a paradigm shift from small kernels (e.g., $3\times3$ or $5\times5$) towards large-kernel convolutions with kernel sizes of $7\times7$ or larger~\cite{liu2022convnet, liu2022more, ding2022scaling, chen2024pelk}. This shift is driven by the need to model long-range dependencies in high-dimensional vision data. Large-kernel convolutions expand the receptive field, capture broader contextual information, and reduce depth redundancy compared to stacks of small kernels. Pioneering architectures like ConvNeXt~\cite{liu2022convnet} and RepLKNet~\cite{ding2022scaling} demonstrate that kernels as large as $31\times31$ can substantially boost performance across diverse vision tasks. However, the quadratic parameter growth relative to kernel size presents significant deployment challenges on resource-constrained devices.

\paragraph{Parameter Compression in Large-Kernel CNNs}

To mitigate the parameter explosion in large-kernel convolutions, SOTA efficient architectures predominantly adopt depthwise separable convolution as their backbone, comprising depthwise convolution and pointwise convolution. Within this framework, significant research effort has focused on compressing the depthwise convolution component. For instance, SLaK~\cite{liu2022more} employs kernel decomposition (e.g., $51\times5 + 5\times51$) and dynamic sparsity to achieve lightweight kernels up to $51\times51$, effectively reducing computational overhead. Similarly, PeLK~\cite{chen2024pelk} utilizes sophisticated parameter sharing strategies, specifically a ``focus-blur" mechanism with exponentially scaled sharing grids and kernel positional embeddings, to support massive kernels up to $101\times101$ while achieving over 90\% parameter reduction in the depthwise convolution. These innovations have successfully made large-kernel models more practical for deployment by tackling the parameter burden of depthwise operations. However, these advances exclusively optimize the depthwise component. As  shown in Fig.~\ref{fig:pw_model}, they overlook the fact that pointwise convolution (1$\times$1 convolution), responsible for crucial channel fusion, becomes the dominant parameter contributor in such models, accounting for over 87\% of total parameters in typical large-kernel CNNs. This represents a major, yet largely unaddressed, barrier to storage-efficient deployment.

\paragraph{Parameter Sharing Strategies}

Parameter sharing is a well-established paradigm for model compression across various architectures~\cite{dong2023efficient}, with specialized implementations emerging for large-kernel CNNs. PeLK~\cite{chen2024pelk}, as mentioned, represents a significant advancement in this domain for depthwise convolution. Its core contribution is a peripheral convolution framework combined with aggressive parameter sharing. By concentrating unique parameters in the kernel center (``focus") and sharing parameters extensively in the periphery (``blur") via an exponentially scaled grid, PeLK~\cite{chen2024pelk} dramatically reduces depthwise convolution parameters while attempting to balance computational efficiency and spatial detail retention, validated on benchmarks like ImageNet-1K classification~\cite{deng2009imagenet} and semantic segmentation on ADE20K~\cite{zhou2019semantic}. While highly effective for depthwise compression, PeLK~\cite{chen2024pelk} and similar strategies do not target the pointwise convolution layer. Consequently, strategies for compressing the parameter-dominant pointwise convolution within large-kernel CNNs remain an open and critical research gap, which our work directly addresses.

\subsection{Model Compression and Deployment on Mobile Devices}

The deployment of deep learning models on resource-constrained mobile devices presents unique challenges that necessitate specialized compression techniques. Traditional model compression methods have evolved to address the stringent memory, computational, and energy constraints inherent to edge devices, with particular emphasis on maintaining inference accuracy while drastically reducing model footprint~\cite{sharma2022enabling}.

Quantization has emerged as a foundational technique for mobile deployment, where 8-bit integer quantization has been shown to achieve near-floating-point accuracy while reducing model size and accelerating inference through specialized hardware instructions~\cite{jacob2018quantization}. Efficient architectures such as MobileNetV2~\cite{sandler2018mobilenetv2} further demonstrate the synergy between lightweight design and quantization. 

Pruning methodologies have similarly undergone significant refinement for mobile scenarios. Structured pruning techniques~\cite{liu2017learning} gained prominence by eliminating entire convolutional filters or channels, thereby producing hardware-friendly models that circumvent the computational inefficiencies associated with sparse matrix operations in unstructured pruning. The synergy between pruning and quantization was particularly impactful, as demonstrated by frameworks that jointly optimized both techniques to achieve 10-20x compression ratios without catastrophic accuracy drops~\cite{han2016deepcompression}.

The practical deployment of these compressed models has been facilitated by mobile-optimized inference engines such as TensorFlow Lite~\cite{david2021tensorflow} and Open Neural Network Exchange (ONNX) Runtime~\cite{onnxruntime}. These frameworks implement critical optimizations including operator fusion, memory-efficient tensor layouts, and hardware-specific kernel implementations, which collectively reduce inference latency by 3-5x compared to naive deployments. Moreover, they provide standardized pathways for deploying models compressed via diverse methodologies, thereby bridging the gap between algorithmic advances and practical implementation~\cite{jiang2020mnn}.

Our CGS method advances this field through a mathematically grounded compression strategy specifically designed for the parameter structure of large-kernel CNNs. Unlike general-purpose techniques, CGS directly targets the parameter-dominant pointwise convolution via an SVD-inspired group-sharing mechanism. This focused approach achieves high compression ratios (e.g., $>$81\% parameter reduction for RepLKNet-31B) while preserving the structural regularity required for efficient mobile inference. Consequently, CGS simultaneously alleviates the critical mobile deployment bottlenecks: stringent storage limits and high memory bandwidth pressure during model loading.
\section{Discussion}

\subsection{Core Design Principles}

The CGS framework is built upon globally shared down/up-projection matrices across all channel groups, enhancing representational consistency and parameter efficiency. This design enforces geometric uniformity in feature maps and reduces the parameter footprint. To capture group-specific variations without sacrificing efficiency, a learnable diagonal scaling matrix is incorporated per channel group. This lightweight component fine-tunes the shared bases, introducing only linear parameter overhead and serving as an efficient alternative to full independent projections per group.

\subsection{Training Stability and Deployment Efficiency}

A notable advantage of CGS lies in its stable optimization dynamics. This property stems from three intrinsic characteristics: the low-rank bottleneck limits basis redundancy, gradient averaging across channel groups mitigates directional noise, and the projection matrices gradually adapt to dominant feature subspaces during training.

From an engineering perspective, the shared architecture is naturally compatible with INT8 quantization, as the shared projections only need to be quantized once and can be reused across groups, drastically reducing overhead and enabling efficient weight reuse on NPUs. The lightweight per-group scaling matrices further preserve accuracy by mitigating quantization error.

Consequently, CGS unifies efficient architectural design, stable training behavior, and hardware-friendly deployment into a practical framework for edge computing.

\subsection{Future Directions}

Several promising directions remain for future research. First, although this work is designed and evaluated for pointwise convolutions in large-kernel CNNs, extending the proposed group-shared low-rank approximation strategy to other matrix-based components, such as Transformer feed-forward networks and attention projections, warrants further investigation.

Second, CGS currently focuses on later network stages, where most parameters reside in the large-kernel CNNs considered in this work. Adapting the method to architectures with more parameter-intensive early layers is another promising direction.

Finally, this work focuses on the widely adopted INT8 quantization setting. Exploring more aggressive quantization schemes, such as INT4 or mixed-precision quantization, and their compatibility with CGS may further improve deployment efficiency.
\section{Conclusions}

This work presents Channel Group-Shared (CGS) low-rank approximation, a novel parameter compression strategy grounded in Singular Value Decomposition (SVD) theory, which effectively resolves critical deployment constraints in modern large-kernel CNNs. Our systematic analysis reveals that pointwise convolutions account for over 87\% of parameters in leading architectures including RepLKNet-31B~\cite{ding2022scaling} and ConvNeXt-B~\cite{liu2022convnet}. This severe parametric imbalance constitutes the fundamental barrier to edge deployment, motivating our development of a geometrically interpretable compression framework. Our method decomposes convolutional kernels into two isomorphic components: (1) shared down/up-projection matrices capturing cross-channel correlations across layer-wide channel groups, analogous to unitary matrices in SVD factorization; (2) lightweight group-specific diagonal scaling matrices preserving channel-adaptive feature transformations, functionally equivalent to singular values.

This design achieves breakthrough parameter efficiency, significantly reducing storage overhead and peak memory load during model initialization, while simultaneously enhancing execution efficiency on mobile devices. Consequently, it enables the successful deployment of large-kernel backbones on resource-constrained mobile devices, overcoming the dual constraints of storage and runtime memory that have hindered edge deployment, while establishing a new paradigm for lightweight network design.

\section{Acknowledgements}

W. Dong's participation was in part supported by the National Natural Science Foundation of China (General Program) (Grant No. 62576267), the Major Science and Technology Innovation Project of Xianyang (Program No. L2025-ZDKJ-ZDGG-ZYRGZN-002), and the Major Science and Technology Innovation Project of Xianyang (Program No. L2024-ZDKJ-ZDGG-GY-0010).
\bibliographystyle{ACM-Reference-Format}
\bibliography{main}


\begin{thebibliography}{67}


\ifx \showCODEN    \undefined \def \showCODEN     #1{\unskip}     \fi
\ifx \showISBNx    \undefined \def \showISBNx     #1{\unskip}     \fi
\ifx \showISBNxiii \undefined \def \showISBNxiii  #1{\unskip}     \fi
\ifx \showISSN     \undefined \def \showISSN      #1{\unskip}     \fi
\ifx \showLCCN     \undefined \def \showLCCN      #1{\unskip}     \fi
\ifx \shownote     \undefined \def \shownote      #1{#1}          \fi
\ifx \showarticletitle \undefined \def \showarticletitle #1{#1}   \fi
\ifx \showURL      \undefined \def \showURL       {\relax}        \fi
\providecommand\bibfield[2]{#2}
\providecommand\bibinfo[2]{#2}
\providecommand\natexlab[1]{#1}
\providecommand\showeprint[2][]{arXiv:#2}

\bibitem[Bello et~al\mbox{.}(2019)]%
        {bello2019attention}
\bibfield{author}{\bibinfo{person}{Irwan Bello}, \bibinfo{person}{Barret Zoph}, \bibinfo{person}{Ashish Vaswani}, \bibinfo{person}{Jonathon Shlens}, {and} \bibinfo{person}{Quoc~V Le}.} \bibinfo{year}{2019}\natexlab{}.
\newblock \showarticletitle{Attention augmented convolutional networks}. In \bibinfo{booktitle}{\emph{Proceedings of the IEEE/CVF international conference on computer vision}}. \bibinfo{pages}{3286--3295}.
\newblock


\bibitem[Cai and Vasconcelos(2018)]%
        {cai2018cascade}
\bibfield{author}{\bibinfo{person}{Zhaowei Cai} {and} \bibinfo{person}{Nuno Vasconcelos}.} \bibinfo{year}{2018}\natexlab{}.
\newblock \showarticletitle{Cascade r-cnn: Delving into high quality object detection}. In \bibinfo{booktitle}{\emph{Proceedings of the IEEE conference on computer vision and pattern recognition}}. \bibinfo{pages}{6154--6162}.
\newblock


\bibitem[Carion et~al\mbox{.}(2020)]%
        {carion2020end}
\bibfield{author}{\bibinfo{person}{Nicolas Carion}, \bibinfo{person}{Francisco Massa}, \bibinfo{person}{Gabriel Synnaeve}, \bibinfo{person}{Nicolas Usunier}, \bibinfo{person}{Alexander Kirillov}, {and} \bibinfo{person}{Sergey Zagoruyko}.} \bibinfo{year}{2020}\natexlab{}.
\newblock \showarticletitle{End-to-end object detection with transformers}. In \bibinfo{booktitle}{\emph{European conference on computer vision}}. Springer, \bibinfo{pages}{213--229}.
\newblock


\bibitem[Chen et~al\mbox{.}(2024)]%
        {chen2024pelk}
\bibfield{author}{\bibinfo{person}{Honghao Chen}, \bibinfo{person}{Xiangxiang Chu}, \bibinfo{person}{Yongjian Ren}, \bibinfo{person}{Xin Zhao}, {and} \bibinfo{person}{Kaiqi Huang}.} \bibinfo{year}{2024}\natexlab{}.
\newblock \showarticletitle{Pelk: Parameter-efficient large kernel convnets with peripheral convolution}. In \bibinfo{booktitle}{\emph{Proceedings of the IEEE/CVF conference on computer vision and pattern recognition}}. \bibinfo{pages}{5557--5567}.
\newblock


\bibitem[Chen et~al\mbox{.}(2019)]%
        {chen2019mmdetection}
\bibfield{author}{\bibinfo{person}{Kai Chen}, \bibinfo{person}{Jiaqi Wang}, \bibinfo{person}{Jiangmiao Pang}, \bibinfo{person}{Yuhang Cao}, \bibinfo{person}{Yu Xiong}, \bibinfo{person}{Xiaoxiao Li}, \bibinfo{person}{Shuyang Sun}, \bibinfo{person}{Wansen Feng}, \bibinfo{person}{Ziwei Liu}, \bibinfo{person}{Jiarui Xu}, {et~al\mbox{.}}} \bibinfo{year}{2019}\natexlab{}.
\newblock \showarticletitle{MMDetection: Open mmlab detection toolbox and benchmark}.
\newblock \bibinfo{journal}{\emph{arXiv preprint arXiv:1906.07155}} (\bibinfo{year}{2019}).
\newblock


\bibitem[Chen et~al\mbox{.}(2014)]%
        {chen2014diannao}
\bibfield{author}{\bibinfo{person}{Tianshi Chen}, \bibinfo{person}{Zidong Du}, \bibinfo{person}{Ninghui Sun}, \bibinfo{person}{Jia Wang}, \bibinfo{person}{Chengyong Wu}, \bibinfo{person}{Yunji Chen}, {and} \bibinfo{person}{Olivier Temam}.} \bibinfo{year}{2014}\natexlab{}.
\newblock \showarticletitle{Diannao: A small-footprint high-throughput accelerator for ubiquitous machine-learning}.
\newblock \bibinfo{journal}{\emph{ACM SIGARCH Computer Architecture News}} \bibinfo{volume}{42}, \bibinfo{number}{1} (\bibinfo{year}{2014}), \bibinfo{pages}{269--284}.
\newblock


\bibitem[Cheng et~al\mbox{.}(2021)]%
        {cheng2021per}
\bibfield{author}{\bibinfo{person}{Bowen Cheng}, \bibinfo{person}{Alex Schwing}, {and} \bibinfo{person}{Alexander Kirillov}.} \bibinfo{year}{2021}\natexlab{}.
\newblock \showarticletitle{Per-pixel classification is not all you need for semantic segmentation}.
\newblock \bibinfo{journal}{\emph{Advances in neural information processing systems}}  \bibinfo{volume}{34} (\bibinfo{year}{2021}), \bibinfo{pages}{17864--17875}.
\newblock


\bibitem[Chollet(2017)]%
        {chollet2017xception}
\bibfield{author}{\bibinfo{person}{Fran{\c{c}}ois Chollet}.} \bibinfo{year}{2017}\natexlab{}.
\newblock \showarticletitle{Xception: Deep learning with depthwise separable convolutions}. In \bibinfo{booktitle}{\emph{Proceedings of the IEEE conference on computer vision and pattern recognition}}. \bibinfo{pages}{1251--1258}.
\newblock


\bibitem[Contributors(2020)]%
        {contributors2020mmsegmentation}
\bibfield{author}{\bibinfo{person}{MMSegmentation Contributors}.} \bibinfo{year}{2020}\natexlab{}.
\newblock \bibinfo{title}{MMSegmentation: Openmmlab semantic segmentation toolbox and benchmark}.
\newblock


\bibitem[Cubuk et~al\mbox{.}(2020)]%
        {cubuk2020randaugment}
\bibfield{author}{\bibinfo{person}{Ekin~D Cubuk}, \bibinfo{person}{Barret Zoph}, \bibinfo{person}{Jonathon Shlens}, {and} \bibinfo{person}{Quoc~V Le}.} \bibinfo{year}{2020}\natexlab{}.
\newblock \showarticletitle{Randaugment: Practical automated data augmentation with a reduced search space}. In \bibinfo{booktitle}{\emph{Proceedings of the IEEE/CVF conference on computer vision and pattern recognition workshops}}. \bibinfo{pages}{702--703}.
\newblock


\bibitem[Dai et~al\mbox{.}(2021)]%
        {dai2021dynamic}
\bibfield{author}{\bibinfo{person}{Xiyang Dai}, \bibinfo{person}{Yinpeng Chen}, \bibinfo{person}{Bin Xiao}, \bibinfo{person}{Dongdong Chen}, \bibinfo{person}{Mengchen Liu}, \bibinfo{person}{Lu Yuan}, {and} \bibinfo{person}{Lei Zhang}.} \bibinfo{year}{2021}\natexlab{}.
\newblock \showarticletitle{Dynamic head: Unifying object detection heads with attentions}. In \bibinfo{booktitle}{\emph{Proceedings of the IEEE/CVF conference on computer vision and pattern recognition}}. \bibinfo{pages}{7373--7382}.
\newblock


\bibitem[David et~al\mbox{.}(2021)]%
        {david2021tensorflow}
\bibfield{author}{\bibinfo{person}{Robert David}, \bibinfo{person}{Jared Duke}, \bibinfo{person}{Advait Jain}, \bibinfo{person}{Vijay~Janapa Reddi}, \bibinfo{person}{Nat Jeffries}, \bibinfo{person}{Jian Li}, \bibinfo{person}{Nick Kreeger}, \bibinfo{person}{Ian Nappier}, \bibinfo{person}{Meghna Natraj}, \bibinfo{person}{Tiezhen Wang}, \bibinfo{person}{Pete Warden}, {and} \bibinfo{person}{Rocky Rhodes}.} \bibinfo{year}{2021}\natexlab{}.
\newblock \showarticletitle{TensorFlow Lite Micro: Embedded Machine Learning for TinyML Systems}. In \bibinfo{booktitle}{\emph{Proceedings of Machine Learning and Systems 3 (MLSys)}}.
\newblock


\bibitem[Deng et~al\mbox{.}(2009)]%
        {deng2009imagenet}
\bibfield{author}{\bibinfo{person}{Jia Deng}, \bibinfo{person}{Wei Dong}, \bibinfo{person}{Richard Socher}, \bibinfo{person}{Li-Jia Li}, \bibinfo{person}{Kai Li}, {and} \bibinfo{person}{Li Fei-Fei}.} \bibinfo{year}{2009}\natexlab{}.
\newblock \showarticletitle{Imagenet: A large-scale hierarchical image database}. In \bibinfo{booktitle}{\emph{2009 IEEE conference on computer vision and pattern recognition}}. Ieee, \bibinfo{pages}{248--255}.
\newblock


\bibitem[developers(2021)]%
        {onnxruntime}
\bibfield{author}{\bibinfo{person}{ONNX~Runtime developers}.} \bibinfo{year}{2021}\natexlab{}.
\newblock \bibinfo{title}{ONNX Runtime}.
\newblock \bibinfo{howpublished}{\url{https://onnxruntime.ai/}}.
\newblock
\newblock
\shownote{Version: x.y.z}.


\bibitem[Ding et~al\mbox{.}(2022)]%
        {ding2022scaling}
\bibfield{author}{\bibinfo{person}{Xiaohan Ding}, \bibinfo{person}{Xiangyu Zhang}, \bibinfo{person}{Jungong Han}, {and} \bibinfo{person}{Guiguang Ding}.} \bibinfo{year}{2022}\natexlab{}.
\newblock \showarticletitle{Scaling up your kernels to 31x31: Revisiting large kernel design in cnns}. In \bibinfo{booktitle}{\emph{Proceedings of the IEEE/CVF conference on computer vision and pattern recognition}}. \bibinfo{pages}{11963--11975}.
\newblock


\bibitem[Dong et~al\mbox{.}(2023)]%
        {dong2023efficient}
\bibfield{author}{\bibinfo{person}{Wei Dong}, \bibinfo{person}{Dawei Yan}, \bibinfo{person}{Zhijun Lin}, {and} \bibinfo{person}{Peng Wang}.} \bibinfo{year}{2023}\natexlab{}.
\newblock \showarticletitle{Efficient adaptation of large vision transformer via adapter re-composing}.
\newblock \bibinfo{journal}{\emph{Advances in Neural Information Processing Systems}}  \bibinfo{volume}{36} (\bibinfo{year}{2023}), \bibinfo{pages}{52548--52567}.
\newblock


\bibitem[Dosovitskiy et~al\mbox{.}(2020)]%
        {dosovitskiy2020image}
\bibfield{author}{\bibinfo{person}{Alexey Dosovitskiy}, \bibinfo{person}{Lucas Beyer}, \bibinfo{person}{Alexander Kolesnikov}, \bibinfo{person}{Dirk Weissenborn}, \bibinfo{person}{Xiaohua Zhai}, \bibinfo{person}{Thomas Unterthiner}, \bibinfo{person}{Mostafa Dehghani}, \bibinfo{person}{Matthias Minderer}, \bibinfo{person}{Georg Heigold}, \bibinfo{person}{Sylvain Gelly}, {et~al\mbox{.}}} \bibinfo{year}{2020}\natexlab{}.
\newblock \showarticletitle{An image is worth 16x16 words: Transformers for image recognition at scale}.
\newblock \bibinfo{journal}{\emph{arXiv preprint arXiv:2010.11929}} (\bibinfo{year}{2020}).
\newblock


\bibitem[Garc{\'\i}a-Ardila et~al\mbox{.}(2018)]%
        {garcia2018canonical}
\bibfield{author}{\bibinfo{person}{Juan~Carlos Garc{\'\i}a-Ardila}, \bibinfo{person}{Luis~E Garza}, {and} \bibinfo{person}{Francisco Marcell{\'a}n}.} \bibinfo{year}{2018}\natexlab{}.
\newblock \showarticletitle{A canonical Geronimus transformation for matrix orthogonal polynomials}.
\newblock \bibinfo{journal}{\emph{Linear and Multilinear Algebra}} \bibinfo{volume}{66}, \bibinfo{number}{2} (\bibinfo{year}{2018}), \bibinfo{pages}{357--381}.
\newblock


\bibitem[Hagos(2018)]%
        {hagos2018android}
\bibfield{author}{\bibinfo{person}{Ted Hagos}.} \bibinfo{year}{2018}\natexlab{}.
\newblock \showarticletitle{Android studio}.
\newblock In \bibinfo{booktitle}{\emph{Learn Android Studio 3: Efficient Android App Development}}. \bibinfo{publisher}{Springer}, \bibinfo{pages}{5--17}.
\newblock


\bibitem[Han et~al\mbox{.}(2016)]%
        {han2016deepcompression}
\bibfield{author}{\bibinfo{person}{Song Han}, \bibinfo{person}{Huizi Mao}, {and} \bibinfo{person}{William~J. Dally}.} \bibinfo{year}{2016}\natexlab{}.
\newblock \showarticletitle{Deep Compression: Compressing Deep Neural Networks with Pruning, Trained Quantization and Huffman Coding}. In \bibinfo{booktitle}{\emph{International Conference on Learning Representations (ICLR)}}.
\newblock
\showeprint[arxiv]{1510.00149}~[cs.CV]
\newblock
\shownote{Published as a conference paper at ICLR 2016 (oral), arXiv:1510.00149 [cs.CV]}.


\bibitem[He et~al\mbox{.}(2017)]%
        {he2017mask}
\bibfield{author}{\bibinfo{person}{Kaiming He}, \bibinfo{person}{Georgia Gkioxari}, \bibinfo{person}{Piotr Doll{\'a}r}, {and} \bibinfo{person}{Ross Girshick}.} \bibinfo{year}{2017}\natexlab{}.
\newblock \showarticletitle{Mask r-cnn}. In \bibinfo{booktitle}{\emph{Proceedings of the IEEE international conference on computer vision}}. \bibinfo{pages}{2961--2969}.
\newblock


\bibitem[He et~al\mbox{.}(2016)]%
        {he2016deep}
\bibfield{author}{\bibinfo{person}{Kaiming He}, \bibinfo{person}{Xiangyu Zhang}, \bibinfo{person}{Shaoqing Ren}, {and} \bibinfo{person}{Jian Sun}.} \bibinfo{year}{2016}\natexlab{}.
\newblock \showarticletitle{Deep residual learning for image recognition}. In \bibinfo{booktitle}{\emph{Proceedings of the IEEE conference on computer vision and pattern recognition}}. \bibinfo{pages}{770--778}.
\newblock


\bibitem[Howard et~al\mbox{.}(2017)]%
        {howard2017mobilenets}
\bibfield{author}{\bibinfo{person}{Andrew~G Howard}, \bibinfo{person}{Menglong Zhu}, \bibinfo{person}{Bo Chen}, \bibinfo{person}{Dmitry Kalenichenko}, \bibinfo{person}{Weijun Wang}, \bibinfo{person}{Tobias Weyand}, \bibinfo{person}{Marco Andreetto}, {and} \bibinfo{person}{Hartwig Adam}.} \bibinfo{year}{2017}\natexlab{}.
\newblock \showarticletitle{Mobilenets: Efficient convolutional neural networks for mobile vision applications}.
\newblock \bibinfo{journal}{\emph{arXiv preprint arXiv:1704.04861}} (\bibinfo{year}{2017}).
\newblock


\bibitem[Hua et~al\mbox{.}(2018)]%
        {hua2018pointwise}
\bibfield{author}{\bibinfo{person}{Binh-Son Hua}, \bibinfo{person}{Minh-Khoi Tran}, {and} \bibinfo{person}{Sai-Kit Yeung}.} \bibinfo{year}{2018}\natexlab{}.
\newblock \showarticletitle{Pointwise convolutional neural networks}. In \bibinfo{booktitle}{\emph{Proceedings of the IEEE conference on computer vision and pattern recognition}}. \bibinfo{pages}{984--993}.
\newblock


\bibitem[Huang et~al\mbox{.}(2019)]%
        {huang2019ccnet}
\bibfield{author}{\bibinfo{person}{Zilong Huang}, \bibinfo{person}{Xinggang Wang}, \bibinfo{person}{Lichao Huang}, \bibinfo{person}{Chang Huang}, \bibinfo{person}{Yunchao Wei}, {and} \bibinfo{person}{Wenyu Liu}.} \bibinfo{year}{2019}\natexlab{}.
\newblock \showarticletitle{Ccnet: Criss-cross attention for semantic segmentation}. In \bibinfo{booktitle}{\emph{Proceedings of the IEEE/CVF international conference on computer vision}}. \bibinfo{pages}{603--612}.
\newblock


\bibitem[Jacob et~al\mbox{.}(2018)]%
        {jacob2018quantization}
\bibfield{author}{\bibinfo{person}{Benoit Jacob}, \bibinfo{person}{Skirmantas Kligys}, \bibinfo{person}{Bo Chen}, \bibinfo{person}{Menglong Zhu}, \bibinfo{person}{Matthew Tang}, \bibinfo{person}{Andrew Howard}, \bibinfo{person}{Hartwig Adam}, {and} \bibinfo{person}{Dmitry Kalenichenko}.} \bibinfo{year}{2018}\natexlab{}.
\newblock \showarticletitle{Quantization and Training of Neural Networks for Efficient Integer-Arithmetic-Only Inference}. In \bibinfo{booktitle}{\emph{Proceedings of the IEEE Conference on Computer Vision and Pattern Recognition (CVPR)}}. \bibinfo{pages}{2704--2713}.
\newblock


\bibitem[Jiang et~al\mbox{.}(2020)]%
        {jiang2020mnn}
\bibfield{author}{\bibinfo{person}{Xiaotang Jiang}, \bibinfo{person}{Huan Wang}, \bibinfo{person}{Yiliu Chen}, \bibinfo{person}{Ziqi Wu}, \bibinfo{person}{Lichuan Wang}, \bibinfo{person}{Bin Zou}, \bibinfo{person}{Yafeng Yang}, \bibinfo{person}{Zongyang Cui}, \bibinfo{person}{Yu Cai}, \bibinfo{person}{Tianhang Yu}, \bibinfo{person}{Chengfei Lyu}, {and} \bibinfo{person}{Zhihua Wu}.} \bibinfo{year}{2020}\natexlab{}.
\newblock \showarticletitle{MNN: A Universal and Efficient Inference Engine}. In \bibinfo{booktitle}{\emph{Proceedings of Machine Learning and Systems 2 (MLSys)}}.
\newblock


\bibitem[Krizhevsky et~al\mbox{.}(2009)]%
        {krizhevsky2009learning}
\bibfield{author}{\bibinfo{person}{Alex Krizhevsky}, \bibinfo{person}{Geoffrey Hinton}, {et~al\mbox{.}}} \bibinfo{year}{2009}\natexlab{}.
\newblock \showarticletitle{Learning multiple layers of features from tiny images}.
\newblock  (\bibinfo{year}{2009}).
\newblock


\bibitem[Lai and Suda(2018)]%
        {lai2018rethinking}
\bibfield{author}{\bibinfo{person}{Liangzhen Lai} {and} \bibinfo{person}{Naveen Suda}.} \bibinfo{year}{2018}\natexlab{}.
\newblock \showarticletitle{Rethinking machine learning development and deployment for edge devices}.
\newblock \bibinfo{journal}{\emph{arXiv preprint arXiv:1806.07846}} (\bibinfo{year}{2018}).
\newblock


\bibitem[LeCun et~al\mbox{.}(1989)]%
        {lecun1989backpropagation}
\bibfield{author}{\bibinfo{person}{Yann LeCun}, \bibinfo{person}{Bernhard Boser}, \bibinfo{person}{John~S Denker}, \bibinfo{person}{Donnie Henderson}, \bibinfo{person}{Richard~E Howard}, \bibinfo{person}{Wayne Hubbard}, {and} \bibinfo{person}{Lawrence~D Jackel}.} \bibinfo{year}{1989}\natexlab{}.
\newblock \showarticletitle{Backpropagation applied to handwritten zip code recognition}.
\newblock \bibinfo{journal}{\emph{Neural computation}} \bibinfo{volume}{1}, \bibinfo{number}{4} (\bibinfo{year}{1989}), \bibinfo{pages}{541--551}.
\newblock


\bibitem[LeCun et~al\mbox{.}(2002)]%
        {lecun2002gradient}
\bibfield{author}{\bibinfo{person}{Yann LeCun}, \bibinfo{person}{L{\'e}on Bottou}, \bibinfo{person}{Yoshua Bengio}, {and} \bibinfo{person}{Patrick Haffner}.} \bibinfo{year}{2002}\natexlab{}.
\newblock \showarticletitle{Gradient-based learning applied to document recognition}.
\newblock \bibinfo{journal}{\emph{Proc. IEEE}} \bibinfo{volume}{86}, \bibinfo{number}{11} (\bibinfo{year}{2002}), \bibinfo{pages}{2278--2324}.
\newblock


\bibitem[Lee et~al\mbox{.}(2020)]%
        {lee2020parameter}
\bibfield{author}{\bibinfo{person}{Sangho Lee}, \bibinfo{person}{Youngjae Yu}, \bibinfo{person}{Gunhee Kim}, \bibinfo{person}{Thomas Breuel}, \bibinfo{person}{Jan Kautz}, {and} \bibinfo{person}{Yale Song}.} \bibinfo{year}{2020}\natexlab{}.
\newblock \showarticletitle{Parameter efficient multimodal transformers for video representation learning}.
\newblock \bibinfo{journal}{\emph{arXiv preprint arXiv:2012.04124}} (\bibinfo{year}{2020}).
\newblock


\bibitem[Lin et~al\mbox{.}(2014)]%
        {lin2014microsoft}
\bibfield{author}{\bibinfo{person}{Tsung-Yi Lin}, \bibinfo{person}{Michael Maire}, \bibinfo{person}{Serge Belongie}, \bibinfo{person}{James Hays}, \bibinfo{person}{Pietro Perona}, \bibinfo{person}{Deva Ramanan}, \bibinfo{person}{Piotr Doll{\'a}r}, {and} \bibinfo{person}{C~Lawrence Zitnick}.} \bibinfo{year}{2014}\natexlab{}.
\newblock \showarticletitle{Microsoft coco: Common objects in context}. In \bibinfo{booktitle}{\emph{European conference on computer vision}}. Springer, \bibinfo{pages}{740--755}.
\newblock


\bibitem[Liu et~al\mbox{.}(2021a)]%
        {liu2021pay}
\bibfield{author}{\bibinfo{person}{Hanxiao Liu}, \bibinfo{person}{Zihang Dai}, \bibinfo{person}{David So}, {and} \bibinfo{person}{Quoc~V Le}.} \bibinfo{year}{2021}\natexlab{a}.
\newblock \showarticletitle{Pay attention to mlps}.
\newblock \bibinfo{journal}{\emph{Advances in neural information processing systems}}  \bibinfo{volume}{34} (\bibinfo{year}{2021}), \bibinfo{pages}{9204--9215}.
\newblock


\bibitem[Liu et~al\mbox{.}(2022a)]%
        {liu2022more}
\bibfield{author}{\bibinfo{person}{Shiwei Liu}, \bibinfo{person}{Tianlong Chen}, \bibinfo{person}{Xiaohan Chen}, \bibinfo{person}{Xuxi Chen}, \bibinfo{person}{Qiao Xiao}, \bibinfo{person}{Boqian Wu}, \bibinfo{person}{Tommi K{\"a}rkk{\"a}inen}, \bibinfo{person}{Mykola Pechenizkiy}, \bibinfo{person}{Decebal Mocanu}, {and} \bibinfo{person}{Zhangyang Wang}.} \bibinfo{year}{2022}\natexlab{a}.
\newblock \showarticletitle{More convnets in the 2020s: Scaling up kernels beyond 51x51 using sparsity}.
\newblock \bibinfo{journal}{\emph{arXiv preprint arXiv:2207.03620}} (\bibinfo{year}{2022}).
\newblock


\bibitem[Liu et~al\mbox{.}(2017)]%
        {liu2017learning}
\bibfield{author}{\bibinfo{person}{Zhuang Liu}, \bibinfo{person}{Jianguo Li}, \bibinfo{person}{Zhiqiang Shen}, \bibinfo{person}{Gao Huang}, \bibinfo{person}{Shoumeng Yan}, {and} \bibinfo{person}{Changshui Zhang}.} \bibinfo{year}{2017}\natexlab{}.
\newblock \showarticletitle{Learning Efficient Convolutional Networks Through Network Slimming}. In \bibinfo{booktitle}{\emph{Proceedings of the IEEE International Conference on Computer Vision (ICCV)}}. \bibinfo{pages}{2736--2744}.
\newblock


\bibitem[Liu et~al\mbox{.}(2021b)]%
        {liu2021swin}
\bibfield{author}{\bibinfo{person}{Ze Liu}, \bibinfo{person}{Yutong Lin}, \bibinfo{person}{Yue Cao}, \bibinfo{person}{Han Hu}, \bibinfo{person}{Yixuan Wei}, \bibinfo{person}{Zheng Zhang}, \bibinfo{person}{Stephen Lin}, {and} \bibinfo{person}{Baining Guo}.} \bibinfo{year}{2021}\natexlab{b}.
\newblock \showarticletitle{Swin transformer: Hierarchical vision transformer using shifted windows}. In \bibinfo{booktitle}{\emph{Proceedings of the IEEE/CVF international conference on computer vision}}. \bibinfo{pages}{10012--10022}.
\newblock


\bibitem[Liu et~al\mbox{.}(2022b)]%
        {liu2022convnet}
\bibfield{author}{\bibinfo{person}{Zhuang Liu}, \bibinfo{person}{Hanzi Mao}, \bibinfo{person}{Chao-Yuan Wu}, \bibinfo{person}{Christoph Feichtenhofer}, \bibinfo{person}{Trevor Darrell}, {and} \bibinfo{person}{Saining Xie}.} \bibinfo{year}{2022}\natexlab{b}.
\newblock \showarticletitle{A convnet for the 2020s}. In \bibinfo{booktitle}{\emph{Proceedings of the IEEE/CVF conference on computer vision and pattern recognition}}. \bibinfo{pages}{11976--11986}.
\newblock


\bibitem[Loshchilov and Hutter(2017)]%
        {loshchilov2017decoupled}
\bibfield{author}{\bibinfo{person}{Ilya Loshchilov} {and} \bibinfo{person}{Frank Hutter}.} \bibinfo{year}{2017}\natexlab{}.
\newblock \showarticletitle{Decoupled weight decay regularization}.
\newblock \bibinfo{journal}{\emph{arXiv preprint arXiv:1711.05101}} (\bibinfo{year}{2017}).
\newblock


\bibitem[Min et~al\mbox{.}(2022)]%
        {min2022peripheral}
\bibfield{author}{\bibinfo{person}{Juhong Min}, \bibinfo{person}{Yucheng Zhao}, \bibinfo{person}{Chong Luo}, {and} \bibinfo{person}{Minsu Cho}.} \bibinfo{year}{2022}\natexlab{}.
\newblock \showarticletitle{Peripheral vision transformer}.
\newblock \bibinfo{journal}{\emph{Advances in Neural Information Processing Systems}}  \bibinfo{volume}{35} (\bibinfo{year}{2022}), \bibinfo{pages}{32097--32111}.
\newblock


\bibitem[Radosavovic et~al\mbox{.}(2020)]%
        {radosavovic2020designing}
\bibfield{author}{\bibinfo{person}{Ilija Radosavovic}, \bibinfo{person}{Raj~Prateek Kosaraju}, \bibinfo{person}{Ross Girshick}, \bibinfo{person}{Kaiming He}, {and} \bibinfo{person}{Piotr Doll{\'a}r}.} \bibinfo{year}{2020}\natexlab{}.
\newblock \showarticletitle{Designing network design spaces}. In \bibinfo{booktitle}{\emph{Proceedings of the IEEE/CVF conference on computer vision and pattern recognition}}. \bibinfo{pages}{10428--10436}.
\newblock


\bibitem[Sandler et~al\mbox{.}(2018)]%
        {sandler2018mobilenetv2}
\bibfield{author}{\bibinfo{person}{Mark Sandler}, \bibinfo{person}{Andrew Howard}, \bibinfo{person}{Menglong Zhu}, \bibinfo{person}{Andrey Zhmoginov}, {and} \bibinfo{person}{Liang-Chieh Chen}.} \bibinfo{year}{2018}\natexlab{}.
\newblock \showarticletitle{MobileNetV2: Inverted Residuals and Linear Bottlenecks}. In \bibinfo{booktitle}{\emph{Proceedings of the IEEE Conference on Computer Vision and Pattern Recognition (CVPR)}}. \bibinfo{pages}{4510--4520}.
\newblock


\bibitem[Schuler et~al\mbox{.}(2022)]%
        {schuler2022grouped}
\bibfield{author}{\bibinfo{person}{Joao Paulo~Schwarz Schuler}, \bibinfo{person}{Santiago Romani}, \bibinfo{person}{Mohamed Abdel-Nasser}, \bibinfo{person}{Hatem Rashwan}, {and} \bibinfo{person}{Domenec Puig}.} \bibinfo{year}{2022}\natexlab{}.
\newblock \showarticletitle{Grouped pointwise convolutions reduce parameters in convolutional neural networks}. In \bibinfo{booktitle}{\emph{Mendel}}, Vol.~\bibinfo{volume}{28}. \bibinfo{pages}{23--31}.
\newblock


\bibitem[Sharma and Sarkar(2022)]%
        {sharma2022enabling}
\bibfield{author}{\bibinfo{person}{Divyasheel Sharma} {and} \bibinfo{person}{Santonu Sarkar}.} \bibinfo{year}{2022}\natexlab{}.
\newblock \showarticletitle{Enabling inference and training of deep learning models for ai applications on iot edge devices}.
\newblock In \bibinfo{booktitle}{\emph{Artificial Intelligence-based Internet of Things Systems}}. \bibinfo{publisher}{Springer}, \bibinfo{pages}{267--283}.
\newblock


\bibitem[Shen et~al\mbox{.}(2015)]%
        {sfm_cvpr15}
\bibfield{author}{\bibinfo{person}{Fumin Shen}, \bibinfo{person}{Chunhua Shen}, \bibinfo{person}{Wei Liu}, {and} \bibinfo{person}{Heng~Tao Shen}.} \bibinfo{year}{2015}\natexlab{}.
\newblock \showarticletitle{Supervised Discrete Hashing}. In \bibinfo{booktitle}{\emph{{IEEE} Conference on Computer Vision and Pattern Recognition}}. \bibinfo{pages}{37--45}.
\newblock


\bibitem[Shen et~al\mbox{.}(2018)]%
        {sfm_tpami2018}
\bibfield{author}{\bibinfo{person}{Fumin Shen}, \bibinfo{person}{Yan Xu}, \bibinfo{person}{Li Liu}, \bibinfo{person}{Yang Yang}, \bibinfo{person}{Zi Huang}, {and} \bibinfo{person}{Heng~Tao Shen}.} \bibinfo{year}{2018}\natexlab{}.
\newblock \showarticletitle{Unsupervised Deep Hashing with Similarity-Adaptive and Discrete Optimization}.
\newblock \bibinfo{journal}{\emph{{IEEE} Transactions on Pattern Analysis and Machine Intelligence}} \bibinfo{volume}{40}, \bibinfo{number}{12} (\bibinfo{year}{2018}), \bibinfo{pages}{3034--3044}.
\newblock


\bibitem[Strang(2012)]%
        {strang2012linear}
\bibfield{author}{\bibinfo{person}{Gilbert Strang}.} \bibinfo{year}{2012}\natexlab{}.
\newblock \bibinfo{booktitle}{\emph{Linear algebra and its applications}}.
\newblock


\bibitem[Szegedy et~al\mbox{.}(2016)]%
        {szegedy2016rethinking}
\bibfield{author}{\bibinfo{person}{Christian Szegedy}, \bibinfo{person}{Vincent Vanhoucke}, \bibinfo{person}{Sergey Ioffe}, \bibinfo{person}{Jon Shlens}, {and} \bibinfo{person}{Zbigniew Wojna}.} \bibinfo{year}{2016}\natexlab{}.
\newblock \showarticletitle{Rethinking the inception architecture for computer vision}. In \bibinfo{booktitle}{\emph{Proceedings of the IEEE conference on computer vision and pattern recognition}}. \bibinfo{pages}{2818--2826}.
\newblock


\bibitem[Tai et~al\mbox{.}(2015)]%
        {tai2015convolutional}
\bibfield{author}{\bibinfo{person}{Cheng Tai}, \bibinfo{person}{Tong Xiao}, \bibinfo{person}{Yi Zhang}, \bibinfo{person}{Xiaogang Wang}, {et~al\mbox{.}}} \bibinfo{year}{2015}\natexlab{}.
\newblock \showarticletitle{Convolutional neural networks with low-rank regularization}.
\newblock \bibinfo{journal}{\emph{arXiv preprint arXiv:1511.06067}} (\bibinfo{year}{2015}).
\newblock


\bibitem[Touvron et~al\mbox{.}(2021)]%
        {touvron2021training}
\bibfield{author}{\bibinfo{person}{Hugo Touvron}, \bibinfo{person}{Matthieu Cord}, \bibinfo{person}{Matthijs Douze}, \bibinfo{person}{Francisco Massa}, \bibinfo{person}{Alexandre Sablayrolles}, {and} \bibinfo{person}{Herv{\'e} J{\'e}gou}.} \bibinfo{year}{2021}\natexlab{}.
\newblock \showarticletitle{Training data-efficient image transformers \& distillation through attention}. In \bibinfo{booktitle}{\emph{International conference on machine learning}}. PMLR, \bibinfo{pages}{10347--10357}.
\newblock


\bibitem[Wang et~al\mbox{.}(2021b)]%
        {wang2021max}
\bibfield{author}{\bibinfo{person}{Huiyu Wang}, \bibinfo{person}{Yukun Zhu}, \bibinfo{person}{Hartwig Adam}, \bibinfo{person}{Alan Yuille}, {and} \bibinfo{person}{Liang-Chieh Chen}.} \bibinfo{year}{2021}\natexlab{b}.
\newblock \showarticletitle{Max-deeplab: End-to-end panoptic segmentation with mask transformers}. In \bibinfo{booktitle}{\emph{Proceedings of the IEEE/CVF conference on computer vision and pattern recognition}}. \bibinfo{pages}{5463--5474}.
\newblock


\bibitem[Wang et~al\mbox{.}(2021a)]%
        {wang2021pyramid}
\bibfield{author}{\bibinfo{person}{Wenhai Wang}, \bibinfo{person}{Enze Xie}, \bibinfo{person}{Xiang Li}, \bibinfo{person}{Deng-Ping Fan}, \bibinfo{person}{Kaitao Song}, \bibinfo{person}{Ding Liang}, \bibinfo{person}{Tong Lu}, \bibinfo{person}{Ping Luo}, {and} \bibinfo{person}{Ling Shao}.} \bibinfo{year}{2021}\natexlab{a}.
\newblock \showarticletitle{Pyramid vision transformer: A versatile backbone for dense prediction without convolutions}. In \bibinfo{booktitle}{\emph{Proceedings of the IEEE/CVF international conference on computer vision}}. \bibinfo{pages}{568--578}.
\newblock


\bibitem[Wang et~al\mbox{.}(2022)]%
        {wang2022pvt}
\bibfield{author}{\bibinfo{person}{Wenhai Wang}, \bibinfo{person}{Enze Xie}, \bibinfo{person}{Xiang Li}, \bibinfo{person}{Deng-Ping Fan}, \bibinfo{person}{Kaitao Song}, \bibinfo{person}{Ding Liang}, \bibinfo{person}{Tong Lu}, \bibinfo{person}{Ping Luo}, {and} \bibinfo{person}{Ling Shao}.} \bibinfo{year}{2022}\natexlab{}.
\newblock \showarticletitle{Pvt v2: Improved baselines with pyramid vision transformer}.
\newblock \bibinfo{journal}{\emph{Computational visual media}} \bibinfo{volume}{8}, \bibinfo{number}{3} (\bibinfo{year}{2022}), \bibinfo{pages}{415--424}.
\newblock


\bibitem[Xiao et~al\mbox{.}(2018)]%
        {xiao2018unified}
\bibfield{author}{\bibinfo{person}{Tete Xiao}, \bibinfo{person}{Yingcheng Liu}, \bibinfo{person}{Bolei Zhou}, \bibinfo{person}{Yuning Jiang}, {and} \bibinfo{person}{Jian Sun}.} \bibinfo{year}{2018}\natexlab{}.
\newblock \showarticletitle{Unified perceptual parsing for scene understanding}. In \bibinfo{booktitle}{\emph{Proceedings of the European conference on computer vision (ECCV)}}. \bibinfo{pages}{418--434}.
\newblock


\bibitem[Xie et~al\mbox{.}(2021)]%
        {xie2021segformer}
\bibfield{author}{\bibinfo{person}{Enze Xie}, \bibinfo{person}{Wenhai Wang}, \bibinfo{person}{Zhiding Yu}, \bibinfo{person}{Anima Anandkumar}, \bibinfo{person}{Jose~M Alvarez}, {and} \bibinfo{person}{Ping Luo}.} \bibinfo{year}{2021}\natexlab{}.
\newblock \showarticletitle{SegFormer: Simple and efficient design for semantic segmentation with transformers}.
\newblock \bibinfo{journal}{\emph{Advances in neural information processing systems}}  \bibinfo{volume}{34} (\bibinfo{year}{2021}), \bibinfo{pages}{12077--12090}.
\newblock


\bibitem[Xie et~al\mbox{.}(2017)]%
        {xie2017aggregated}
\bibfield{author}{\bibinfo{person}{Saining Xie}, \bibinfo{person}{Ross Girshick}, \bibinfo{person}{Piotr Doll{\'a}r}, \bibinfo{person}{Zhuowen Tu}, {and} \bibinfo{person}{Kaiming He}.} \bibinfo{year}{2017}\natexlab{}.
\newblock \showarticletitle{Aggregated residual transformations for deep neural networks}. In \bibinfo{booktitle}{\emph{Proceedings of the IEEE conference on computer vision and pattern recognition}}. \bibinfo{pages}{1492--1500}.
\newblock


\bibitem[Yu and Koltun(2015)]%
        {yu2015multi}
\bibfield{author}{\bibinfo{person}{Fisher Yu} {and} \bibinfo{person}{Vladlen Koltun}.} \bibinfo{year}{2015}\natexlab{}.
\newblock \showarticletitle{Multi-scale context aggregation by dilated convolutions}.
\newblock \bibinfo{journal}{\emph{arXiv preprint arXiv:1511.07122}} (\bibinfo{year}{2015}).
\newblock


\bibitem[Yu et~al\mbox{.}(2017)]%
        {yu2017dilated}
\bibfield{author}{\bibinfo{person}{Fisher Yu}, \bibinfo{person}{Vladlen Koltun}, {and} \bibinfo{person}{Thomas Funkhouser}.} \bibinfo{year}{2017}\natexlab{}.
\newblock \showarticletitle{Dilated residual networks}. In \bibinfo{booktitle}{\emph{Proceedings of the IEEE conference on computer vision and pattern recognition}}. \bibinfo{pages}{472--480}.
\newblock


\bibitem[Yu et~al\mbox{.}(2022)]%
        {yu2022metaformer}
\bibfield{author}{\bibinfo{person}{Weihao Yu}, \bibinfo{person}{Mi Luo}, \bibinfo{person}{Pan Zhou}, \bibinfo{person}{Chenyang Si}, \bibinfo{person}{Yichen Zhou}, \bibinfo{person}{Xinchao Wang}, \bibinfo{person}{Jiashi Feng}, {and} \bibinfo{person}{Shuicheng Yan}.} \bibinfo{year}{2022}\natexlab{}.
\newblock \showarticletitle{Metaformer is actually what you need for vision}. In \bibinfo{booktitle}{\emph{Proceedings of the IEEE/CVF conference on computer vision and pattern recognition}}. \bibinfo{pages}{10819--10829}.
\newblock


\bibitem[Yuan et~al\mbox{.}(2021)]%
        {yuan2021tokens}
\bibfield{author}{\bibinfo{person}{Li Yuan}, \bibinfo{person}{Yunpeng Chen}, \bibinfo{person}{Tao Wang}, \bibinfo{person}{Weihao Yu}, \bibinfo{person}{Yujun Shi}, \bibinfo{person}{Zi-Hang Jiang}, \bibinfo{person}{Francis~EH Tay}, \bibinfo{person}{Jiashi Feng}, {and} \bibinfo{person}{Shuicheng Yan}.} \bibinfo{year}{2021}\natexlab{}.
\newblock \showarticletitle{Tokens-to-token vit: Training vision transformers from scratch on imagenet}. In \bibinfo{booktitle}{\emph{Proceedings of the IEEE/CVF international conference on computer vision}}. \bibinfo{pages}{558--567}.
\newblock


\bibitem[Yuan et~al\mbox{.}(2022)]%
        {yuan2022volo}
\bibfield{author}{\bibinfo{person}{Li Yuan}, \bibinfo{person}{Qibin Hou}, \bibinfo{person}{Zihang Jiang}, \bibinfo{person}{Jiashi Feng}, {and} \bibinfo{person}{Shuicheng Yan}.} \bibinfo{year}{2022}\natexlab{}.
\newblock \showarticletitle{Volo: Vision outlooker for visual recognition}.
\newblock \bibinfo{journal}{\emph{IEEE transactions on pattern analysis and machine intelligence}} \bibinfo{volume}{45}, \bibinfo{number}{5} (\bibinfo{year}{2022}), \bibinfo{pages}{6575--6586}.
\newblock


\bibitem[Yun et~al\mbox{.}(2019)]%
        {yun2019cutmix}
\bibfield{author}{\bibinfo{person}{Sangdoo Yun}, \bibinfo{person}{Dongyoon Han}, \bibinfo{person}{Seong~Joon Oh}, \bibinfo{person}{Sanghyuk Chun}, \bibinfo{person}{Junsuk Choe}, {and} \bibinfo{person}{Youngjoon Yoo}.} \bibinfo{year}{2019}\natexlab{}.
\newblock \showarticletitle{Cutmix: Regularization strategy to train strong classifiers with localizable features}. In \bibinfo{booktitle}{\emph{Proceedings of the IEEE/CVF international conference on computer vision}}. \bibinfo{pages}{6023--6032}.
\newblock


\bibitem[Zhang et~al\mbox{.}(2017)]%
        {zhang2017mixup}
\bibfield{author}{\bibinfo{person}{Hongyi Zhang}, \bibinfo{person}{Moustapha Cisse}, \bibinfo{person}{Yann~N Dauphin}, {and} \bibinfo{person}{David Lopez-Paz}.} \bibinfo{year}{2017}\natexlab{}.
\newblock \showarticletitle{mixup: Beyond empirical risk minimization}.
\newblock \bibinfo{journal}{\emph{arXiv preprint arXiv:1710.09412}} (\bibinfo{year}{2017}).
\newblock


\bibitem[Zhang et~al\mbox{.}(2023)]%
        {zhang2023speeding}
\bibfield{author}{\bibinfo{person}{Meng Zhang}, \bibinfo{person}{Fei Liu}, {and} \bibinfo{person}{Dongpeng Weng}.} \bibinfo{year}{2023}\natexlab{}.
\newblock \showarticletitle{Speeding-up and compression convolutional neural networks by low-rank decomposition without fine-tuning}.
\newblock \bibinfo{journal}{\emph{Journal of Real-Time Image Processing}} \bibinfo{volume}{20}, \bibinfo{number}{4} (\bibinfo{year}{2023}), \bibinfo{pages}{64}.
\newblock


\bibitem[Zhang et~al\mbox{.}(2018)]%
        {zhang2018systematic}
\bibfield{author}{\bibinfo{person}{Tianyun Zhang}, \bibinfo{person}{Shaokai Ye}, \bibinfo{person}{Kaiqi Zhang}, \bibinfo{person}{Jian Tang}, \bibinfo{person}{Wujie Wen}, \bibinfo{person}{Makan Fardad}, {and} \bibinfo{person}{Yanzhi Wang}.} \bibinfo{year}{2018}\natexlab{}.
\newblock \showarticletitle{A systematic dnn weight pruning framework using alternating direction method of multipliers}. In \bibinfo{booktitle}{\emph{Proceedings of the European conference on computer vision (ECCV)}}. \bibinfo{pages}{184--199}.
\newblock


\bibitem[Zhong et~al\mbox{.}(2020)]%
        {zhong2020random}
\bibfield{author}{\bibinfo{person}{Zhun Zhong}, \bibinfo{person}{Liang Zheng}, \bibinfo{person}{Guoliang Kang}, \bibinfo{person}{Shaozi Li}, {and} \bibinfo{person}{Yi Yang}.} \bibinfo{year}{2020}\natexlab{}.
\newblock \showarticletitle{Random erasing data augmentation}. In \bibinfo{booktitle}{\emph{Proceedings of the AAAI conference on artificial intelligence}}, Vol.~\bibinfo{volume}{34}. \bibinfo{pages}{13001--13008}.
\newblock


\bibitem[Zhou et~al\mbox{.}(2019)]%
        {zhou2019semantic}
\bibfield{author}{\bibinfo{person}{Bolei Zhou}, \bibinfo{person}{Hang Zhao}, \bibinfo{person}{Xavier Puig}, \bibinfo{person}{Tete Xiao}, \bibinfo{person}{Sanja Fidler}, \bibinfo{person}{Adela Barriuso}, {and} \bibinfo{person}{Antonio Torralba}.} \bibinfo{year}{2019}\natexlab{}.
\newblock \showarticletitle{Semantic understanding of scenes through the ade20k dataset}.
\newblock \bibinfo{journal}{\emph{International Journal of Computer Vision}}  \bibinfo{volume}{127} (\bibinfo{year}{2019}), \bibinfo{pages}{302--321}.
\newblock


\end{thebibliography}

\appendix

\section{Training Configurations}
\label{Training_Configurations}

\subsection{ImageNet-1K}

To pretrain a classification model on the ImageNet-1K dataset, we employed 8 NVIDIA A800 GPUs and optimized the model using AdamW~\cite{loshchilov2017decoupled} with a weight decay of 0.05. The learning rate was set to 8e-3, accompanied by a 10-epoch linear warmup and a cosine decay schedule. We conducted experiments with 120 and 300 training epochs, where the 120-epoch training was primarily used to observe phenomena in the main and ablation studies. For data augmentation, we used the parameter combination RandAugment~\cite{cubuk2020randaugment} `rand-m9-mstd0.5-inc1', along with label smoothing~\cite{szegedy2016rethinking} (coefficient = 0.1), Mixup~\cite{zhang2017mixup} ($\alpha$ = 0.8), CutMix~\cite{yun2019cutmix} ($\alpha$ = 1.0), and random erasing~\cite{zhong2020random} (probability p = 0.25). The layer scale was initialized to 1e-6, and the Exponential Moving Average (EMA) decay factor was configured as 0.9999.

\subsection{MS-COCO}

We followed the training configuration used in RepLKNet~\cite{ding2022scaling}, and fine-tuned the pre-trained model CGS-B via Cascade Mask R-CNN\cite{cai2018cascade,he2017mask} in  MMdetection~\cite{chen2019mmdetection} with 36 epochs. The hyperparameters used in the object detection process are consistent with those in the Github repository of RepLKNet-pytorch.

\subsection{ADE20K}

We followed the training configuration used in RepLKNet~\cite{ding2022scaling}, and fine-tuned the pre-trained model CGS-B via UperNet~\cite{xiao2018unified} in MMSegmentation~\cite{contributors2020mmsegmentation} with 160K iterations. The hyperparameters used in the semantic segmentation process are consistent with those in the Github repository of RepLKNet-pytorch.

\subsection{CIFAR-100}

We retain the identical training configuration from ImageNet-1K. For ablation analysis, we conduct 200-epoch experiments to observe critical patterns.

\begin{table}[!tb]
\centering
\caption{CIFAR-100 classification accuracy (\%) comparison. For CGS-S/B/L variants, we systematically evaluate the impact of bottleneck dimensions in down/up-projection matrices by varying their sizes, employing RepLKNet as the backbone architecture.}
\begin{tabular}{l|c|c|c}
    \toprule
    Model & \makecell{Bottleneck \\ Dimension} & \makecell{Params \\ (M)} & \makecell{Top-1 \\ Acc} \\
    \midrule
    \multirow{4}{*}{RepLkNet+CGS-S} & 50 & 6.4 & 73.1 \\
     & 100 & 8.7 & 77.3 \\
     & 200 & 13.2 & 79.1 \\
     & 300 & 17.7 & 78.8 \\
    \midrule
    \multirow{4}{*}{RepLkNet+CGS-B} & 50 & 6.9 & 78.4 \\
     & 100 & 9.2 & 80.1 \\
     & 200 & 13.7 & 80.3 \\
     & 300 & 18.3 & 79.2 \\
    \midrule
    \multirow{4}{*}{RepLkNet+CGS-L} & 50 & 9.3 & 78.5 \\
     & 100 & 11.6 & 81.3 \\
     & 200 & 16.2 & 81.6 \\
     & 300 & 20.7 & 82.1 \\
    \bottomrule
\end{tabular}
\label{tab:Ablation_results}
\end{table}

\begin{table}[tbhp]
\centering
\caption{The processor models corresponding to the devices.}
\begin{tabular}{l|c}
    \toprule
    Device & Processor\\
    \midrule
    Google Pixel 2 (Emulator) & Snapdragon 835 \\
    Galaxy Nexus (Emulator) & OMAP4460 \\
    HUAWEI P20 pro & HiSilicon Kirin 970 \\
    HUAWEI nova 8 & Hisilicon Kirin 985 \\
    Honor 200 Pro & Snapdragon 8s Gen 3 \\
    OPPO Reno 4 & Snapdragon 765G \\
    OnePlus 13 & Snapdragon 8 \\
    Xiaomi 14 Ultra & Snapdragon 8 Gen3 \\
    vivo X300 Pro & Dimensity 9500 \\
    \bottomrule
\end{tabular}
\label{tab:processor}
\end{table}

\balance

\section{Detailed Results on Bottleneck Dimensions}
\label{app:Ablation_study_dim}

This section presents a comprehensive empirical evaluation of the down/up-projection matrices in CGS-S/B/L across varying bottleneck dimensions on the CIFAR-100~\cite{krizhevsky2009learning} dataset, encompassing both model performance and parameter count. Tab.~\ref{tab:Ablation_results} systematically quantifies the impact of the bottleneck dimension within the down/up-projection matrices on model performance under the CGS-S/B/L configurations. Experiments were conducted using RepLKNet~\cite{ding2022scaling} as the backbone architecture, with controlled comparisons across bottleneck dimensions of $\{50, 100, 200, 300\}$.

Notably, configurations employing dimensionalities below this critical threshold of 200 exhibit a consistent and significant degradation in model efficacy, failing to achieve the necessary representational capacity. Conversely, increasing the dimensionality beyond 200 incurs substantial and unnecessary costs. This escalation not only induces a substantial increase in parameter overhead, exemplified by a $33.6\%$ parameter count increase observed for the CGS-B configuration when moving from dimension 200 to 300, but also paradoxically leads to a measurable detriment in model accuracy. For instance, the same dimensional increase for CGS-B resulted in a notable accuracy reduction of $1.4\%$. This counter-intuitive performance decline at higher dimensions suggests potential over-parameterization or optimization challenges within the compressed structure. 

Consequently, the dimension of 200 is identified as the singular configuration that simultaneously maximizes accuracy while minimizing unnecessary parameter growth across the evaluated CGS variants.

Furthermore, the parameter increase associated with our proposed CGS-L configuration is marginal compared to CGS-B and CGS-S (e.g., merely +2.5M parameters versus CGS-B at bottleneck dimension 200). This lower parameter cost alleviates concerns regarding scalability when deploying our larger-scale approach.

\section{Detailed Results on Structural and Quantization Synergy}
\label{app:deploy_struc_quant}

\begin{table}[t]
\centering
\caption{Performance comparison on RepLKNet-31B with ImageNet-1K evaluated on a Snapdragon 8 Gen 3 mobile platform.}
\label{tab:real_world_performance}
\begin{tabular}{l|c|c|c|c|c}
    \toprule
    Method & \makecell{Params \\ (M)} & \makecell{FLOPs \\ (G)} & \makecell{Top-1 \\ Acc.} & \makecell{Lat. \\ (ms)} & \makecell{Energy \\ (mWh)} \\
    \midrule
    \makecell{INT8 \\ Quant.~\cite{jacob2018quantization}}      & 79.0 & 15.6 & 82.9 & 265 & 21.5 \\
    \makecell{ADMM \\ Prune~\cite{zhang2018systematic}}       & 42.5 & 9.1  & 82.3 & 290 & 24.0 \\
    \makecell{CGS-B \\ (ours)}     & 14.7 & 28.7 & 80.0 & 198 & 12.2 \\
    \makecell{CGS-B + \\ INT8~\cite{jacob2018quantization}}     & 14.7 & 28.7 & 79.6 & 135 & 7.8  \\
    \bottomrule
\end{tabular}
\end{table}

Tab.~\ref{tab:real_world_performance} compares the real-world performance of various compression methods applied to RepLKNet-31B. The results confirm that CGS effectively optimizes large-kernel CNNs for resource-constrained deployment. Energy consumption denotes the total energy measured over multiple runs.

CGS-B achieves a superior balance of parameters, latency, and energy efficiency compared to standard INT8 quantization and ADMM-based pruning. Although CGS-B has higher FLOPs, its inference latency is significantly lower. This gain stems from improved memory access patterns and better hardware utilization, alleviating the memory-bandwidth bottleneck common in mobile NPUs/GPUs.

CGS-B reduces model size by over 80\% versus the original backbone while maintaining competitive accuracy. In contrast, pruning often creates irregular sparsity, limiting practical speedups. When combined with INT8 quantization, CGS-B further reduces latency and energy by over 30\% with lower accuracy loss, demonstrating strong compatibility with precision reduction.

In summary, CGS enables joint structural and numerical optimization, making it suitable for memory- and power-constrained mobile platforms. These results underscore the value of hardware-aware structural redesign beyond conventional compression.

\section{Mobile Processor Model}
\label{Mobile}

The mobile device processors we used to deploy models and perform inference are shown in Tab.~\ref{tab:processor}.

\section{Code}

Our code is available at \url{https://github.com/LforikCyzzz/CGS_code}.

\end{document}